\pdfoutput=1
\ifdefined\pdfminorversion\pdfminorversion=7\fi
\documentclass{article}
\usepackage{style,times}
\usepackage[T1]{fontenc}

\usepackage{amsmath,amsfonts,bm}

\def\eqref#1{equation~\ref{#1}}

\def\1{\bm{1}}

\DeclareMathAlphabet{\mathsfit}{\encodingdefault}{\sfdefault}{m}{sl}
\SetMathAlphabet{\mathsfit}{bold}{\encodingdefault}{\sfdefault}{bx}{n}

\DeclareMathOperator*{\argmax}{arg\,max}
\DeclareMathOperator*{\argmin}{arg\,min}

\usepackage[hidelinks]{hyperref}
\usepackage{url}
\usepackage{bbm}
\DeclareFontFamily{U}{bbm}{}
\DeclareFontShape{U}{bbm}{m}{n}{<-6> bbm5 <6-7> bbm6 <7-8> bbm7 <8-9> bbm8 <9-10> bbm9 <10-12> bbm10 <12-> bbm12}{}
\usepackage{multirow}
\usepackage{color}
\usepackage{mathtools}
\usepackage{makecell}
\usepackage{lscape}
\usepackage{printlen}
\usepackage{amssymb}
\usepackage{booktabs}
\usepackage{amsfonts}
\usepackage{graphicx}

\title{{\Large Autonomous Learning From Success and Failure:\\ Goal-Conditioned Supervised Learning with Negative Feedback}}

\DeclareMathSymbol{\shortminus}{\mathbin}{AMSa}{"39}

\author{Zeqiang Zhang, Fabian Wurzberger, Gerrit Schmid, Sebastian Gottwald \& Daniel A. Braun  \\
Institute of Neural Information Processing\\
Ulm University\\
Ulm, Germany\\
}

\renewcommand{\bibfont}{\raggedright}

\stylefinalcopy
\begin{document}

\maketitle

\begin{abstract}
Learning from reward functions and imitation learning of
human-generated demonstrations are the two principal approaches for training
autonomous systems that interact with an environment through action and
observation. Both, however, require human specification for each behaviour to be
acquired, a problem for long-lived self-adaptive systems whose goals and operating
conditions cannot be fully anticipated at design time. Recently, Goal-Conditioned
Supervised Learning (GCSL) through self-imitation has been proposed as a
self-supervised alternative: by strategically relabelling goals, agents can derive
policy insights from their own experiences. Despite its successes, this framework
presents two notable limitations: (1) learning exclusively from self-generated
experiences can exacerbate the agents' inherent biases; (2) the relabelling
strategy allows agents to focus solely on successful outcomes, precluding them
from learning from their mistakes. To address these issues, we propose
Goal-Conditioned Supervised Learning with Negative Feedback (GCSL-NF), which
evaluates each trajectory twice: positively with respect to relabelled goals, and
correctively with respect to the goal originally intended. The corrective target
comes from a similarity function learned contrastively from trajectory-induced
neighbourhood relations, so that neither a reward function nor a geometric
distance needs to be specified. Our experiments show that GCSL-NF overcomes
limitations imposed by agents' initial biases, increasingly benefits from negative
feedback as learning progresses, and matches or surpasses leading GCSL- and
HER-based methods across discrete and continuous goal-reaching tasks. Its learning
signal further remains informative under corrupted priors, runtime goal shift, sensor and actuation noise, and constraints enforced only during training. By reducing reliance on prespecified reward functions, the proposed approach is particularly relevant for self-adaptive autonomous systems, where adaptation objectives may be diverse, changing, or difficult to engineer.
\end{abstract}

\section{Introduction}
\label{sec:introduction}

{
Consider an autonomous space robot dispatched to a distant planet with unknown terrain and hazards to deal with the unexpected.  Neither the environment it will face nor all the goals worth pursuing there can be specified before launch, and communication is too slow for a human in the loop. 
Instead, the robot must be able to learn goal-directed behaviour from its own 
experience.
For such a system, a learning mechanism must extract supervision from every trajectory, successful or not, without a prespecified reward function.

This scenario is an instance of a long-standing aim of autonomous learning research: enabling adaptive systems to acquire new 
goal-directed behaviour from their own interactions, without requiring 
hand-engineered objective functions or human-generated demonstrations for every goal they might pursue. This challenge is especially acute for lifelong learning systems, including long-lived autonomous and self-adaptive systems, which must adapt their behaviour at runtime in response to changing goals, environments, and system conditions that cannot be fully anticipated at design time \citep{kephart2003vision, cheng2009software, delemos2013software, bencomo2010requirements,weyns2020software}. 
Self-adaptive systems research often frames adaptation in terms of feedback-loop architectures  with design-time specified adaptation logic \citep{kephart2003vision,aastrom1995adaptive}, and more recently through reinforcement learning with pre-specified reward functions for runtime decision-making~\citep{feit2022explaining}.
Here we focus on versatile open-ended systems that may require an even broader form of behavioral adaptation: the ability to learn and reuse policies for diverse goals that may not have been exhaustively specified at design time.

Within the classical paradigm of reinforcement learning (RL) based on scalar 
rewards~\citep{mnih2015human, silver2017mastering, vinyals2019grandmaster}, 
open-ended learning beyond prespecified tasks is often studied in the context of intrinsic reward functions, 
such as empowerment \citep{empowerment}, learning progress \citep{Oudeyer2007}, or curiosity-driven measures based on the prediction error of a learned forward model \citep{stadie2015incentivizing} or inverse model \citep{Curiosity2017}, and random network distillation \citep{burda2018exploration}. 
A complementary line of work studies goal-based agents that autonomously generate and pursue goals 
based on their own prior experience~\citep{kaelbling1993learning,Oudeyer2022imgep}. 
Goal-based approaches are especially attractive for autonomous and self-adaptive systems because goals can often be expressed directly in state or observation space and varied without redefining a complete 
task-specific reward function. 
By contrast, behaviours induced by intrinsic rewards are not always straightforward to align with 
externally relevant objectives. 
Nevertheless, goal-conditioned RL does not by itself eliminate the reward-specification problem: it often 
still requires a reward or distance function that evaluates progress toward a goal. 
Such predefined distance-based rewards can be domain-dependent and idiosyncratic, and in many environments the resulting feedback 
is sparse or absent for most goals the agent encounters, making learning inefficient
~\citep{ladosz2022exploration, vecerik2017leveraging, 10018434, kaelbling1993learning, andrychowicz2017hindsight}.

Goal-relabelling strategies offer a partial solution by reinterpreting 
the agent's own experience as supervisory signal. Hindsight Experience 
Replay (HER)~\citep{andrychowicz2017hindsight} recomputes rewards under 
relabelled goals and updates policies via off-policy RL. Goal-Conditioned 
Supervised Learning (GCSL)~\citep{ghosh2021learning} goes one step further: 
by treating relabelled trajectories as expert demonstrations, it casts the 
goal-conditioned policy learning problem entirely as supervised learning, 
inheriting the stability and convergence properties of behavioural cloning 
while requiring no external reward function. This makes GCSL well suited to autonomous learning: the agent improves through 
self-imitation of its own trajectories, with goals supplied implicitly by 
the relabelling procedure. However, learning exclusively by self-imitation introduces a structural 
limitation: trajectories that fail to reach their intended goal are 
silently reinterpreted as successful trajectories for some other goal, 
and the information they carry about the \emph{original} goal is 
discarded. The agent has no mechanism for recognising or correcting 
behavioural commitments that are inconsistent with what it was actually 
asked to do. Weighted variants such as W-GCSL~\citep{yang2022rethinking} 
and OCBC~\citep{eysenbach2022imitating} adjust the weights of relabelled 
trajectories but do not introduce negative feedback. As a result, these 
methods remain vulnerable to biased initial policies, can converge to 
suboptimal solutions in the presence of obstacles or non-trivial geometry, 
and offer no signal for the agent to learn from its mistakes.
For the rover above, this means that early wandering habits are relabelled 
as successes and self-reinforced, while the information carried by every 
costly failed drive is discarded.

The core difficulty in developing such a system in an 
autonomous setting is that recovering negative feedback requires the agent to judge, without external 
supervision, whether an achieved final state is close to the intended 
goal. Predefined distance functions in observation space resolve this 
trivially but reintroduce the very reward-engineering bottleneck that 
self-imitation was meant to circumvent: they fail in settings such as 
LiDAR-based observation, where geometric distance carries no reliable 
information about task progress. We address this by learning a distance 
function directly from trajectory data, using contrastive principles to 
estimate whether two states are temporally proximate in the agent's own 
experience. This learned distance turns failed trajectories into a usable 
source of corrective signal, without external reward and without 
geometric assumptions about the observation space.

We call the resulting method \emph{Goal-Conditioned Supervised Learning 
with Negative Feedback (GCSL-NF)}. The method combines two loss terms 
acting on the same policy: a positive term that imitates relabelled 
trajectories (as in standard GCSL) and a negative term that penalizes 
actions leading to states judged distant from the original goal by the 
learned similarity (inverse distance) function. To our knowledge, this is the first method to exploit original-goal failure information within a purely supervised goal-conditioned framework, without any external reward, solely based on a learned similarity function. Our experiments show that GCSL-NF overcomes 
the behavioural biases that limit standard GCSL, surpasses leading 
GCSL-based and HER-based methods across multiple challenging environments. Our method confers measurable robustness to perturbations characteristic of 
deployed autonomous and adaptive systems, including corrupted initial 
policies, sensor and actuation noise, runtime relaxation of training-time 
safety constraints, and history-dependent mission rules 
(Section~\ref{sec:robustness}). The design rationale distinguishing our approach from reinforcement learning on a learned reward and from undirected contrastive negatives is developed in Section~\ref{sec:rw-methods}. The relationship between our contribution 
and the broader literature on autonomous and adaptive systems is 
developed in Section~\ref{sec:positioning}.

}

{
\section{Preliminaries}

\subsection{Problem setting: goal-conditioned learning}
\label{subsec:problem-setting}

Goal-conditioned learning provides a very general framework to formalize adaptive system behavior ranging from simple coordinate-based navigation to complex, vision-guided robotic manipulation and digital UI automation.
We consider finite-horizon goal-conditioned control problems in which an agent must learn to reach goals from its own experience. 
The environment is modelled as a goal-conditioned Markov decision process
$(\mathcal{S},\mathcal{A},\mathcal{G},\mathcal{T}, r_g, \gamma, T)$, where $\mathcal{S}$ is the state space, $\mathcal{A}$ is the action space, $\mathcal{G}$ is the goal space, $\mathcal{T}(s'|s,a)$ is the transition dynamics, $r_g$ is a goal-dependent reward, $\gamma$ is the discount factor, and $T$ is the horizon. 
At the beginning of each episode, a goal $g \sim p(g)$ is sampled, and the agent executes a goal-conditioned policy $\pi(a|s,g)$ to generate a trajectory
$\tau=(s_0,a_0,\ldots,s_T)$. 

In this paper, goals are represented as desired states or observations, i.e.,
$\mathcal{G}\subseteq\mathcal{S}$, or more generally as elements that can be compared to states in the observation space. 
Thus, a goal specifies an intended outcome of the trajectory rather than an external task label. 
This formulation is in line with the majority of the literature \citep{ghosh2021learning, andrychowicz2017hindsight} and covers the sparse goal-reaching settings considered in our experiments, where the agent receives little or no useful reward unless the achieved state is close to the desired goal. 
It also clarifies the scope of the present method: abstract goals, temporal-logic specifications, and hierarchical mission objectives are not assumed to be directly available as goals unless they can be mapped to observable states, waypoints, or rule-checkable trajectory properties. 

We distinguish three goal-related quantities that will be used throughout the paper:
\begin{itemize}
    \item The \emph{original goal} $g$ is the goal sampled before rollout and represents what the agent intended to achieve. 
    \item The \emph{achieved outcome} is the final or future state actually produced by the trajectory, such as $s_T$ or $s_i$. 
    \item A \emph{relabelled goal} $g'$ is an achieved future state from the same trajectory, typically $g'=s_i$ for $i\geq t$, which allows the trajectory to be reused as a successful demonstration for what it did achieve. 
\end{itemize}
The above distinctions are central to GCSL-NF: relabelled goals provide positive hindsight supervision, whereas the mismatch between the achieved outcome and the original goal provides the basis for negative feedback.

\subsection{Hindsight relabelling and goal-conditioned supervised learning}
\label{subsec:gcsl}

A common strategy for sparse goal-conditioned problems is hindsight relabelling. 
Instead of treating a failed trajectory as useless, one can reinterpret the achieved states as goals that the trajectory successfully reached. 
In Hindsight Experience Replay (HER), the reward is recomputed with respect to relabelled goals and the policy is then updated using an off-policy reinforcement learning algorithm. 
By contrast, goal-conditioned supervised learning (GCSL) removes the reward-maximisation step and directly imitates relabelled trajectories.

In GCSL, the training dataset consists of tuples
\[
    \mathcal{D}=\{(s_t,a_t,g')\},
\]
where $g'$ is either an expert-provided goal or a hindsight relabelled goal, typically a future state $s_i$ with $i\geq t$. 
The policy is trained by maximising the likelihood of the recorded action under the relabelled goal:
\[
    \pi = \arg\max_{\pi}
    \mathbb{E}_{(s_t,a_t,g')\sim\mathcal{D}}
    \left[\log \pi(a_t|s_t,g')\right].
\]

This objective uses hindsight relabelling purely as positive supervision. 
If a trajectory was generated with the original goal $g$ but ended at an achieved state $s_T$ far from $g$, GCSL does not explicitly use this mismatch as evidence that the selected actions were unsuitable for reaching $g$. 
Instead, the same trajectory is only reinterpreted as a successful demonstration for some relabelled goal $g'$. 
This is the central limitation addressed by GCSL-NF: the agent should learn not only what its actions succeeded in achieving, but also what they failed to achieve relative to the original goal.

\subsection{Goal attainment, learned similarity, and negative feedback}
\label{subsec:goal-attainment}

Learning from failed attempts requires a criterion for judging whether the originally intended goal was reached. 
In discrete goal-reaching tasks, this can be expressed by an indicator such as $r_g(s)=\mathbbm{1}(s=g)$. 
In continuous or unstructured observation spaces, however, exact equality is usually inappropriate, and goal attainment must instead be assessed through a similarity or distance measure between the achieved state and the intended goal.

Ideally, such a measure would reflect a policy-induced distance: two states should be considered close if the agent can move between them in a small number of steps under an appropriate policy. 
This would provide a task-relevant notion of distance, rather than relying on Euclidean proximity in the raw observation space. 
In general, however, the optimal policy and the corresponding policy-induced distance are not known in advance. 
GCSL-NF therefore learns a similarity function $p_\varphi(s,s')$ from trajectory-induced neighbourhood relations. 
The value $p_\varphi(s,s')$ is interpreted as an estimate of whether two states are close in the transition structure experienced by the agent.

This learned similarity plays two roles in our framework. 
First, it provides an autonomous criterion for approximate goal attainment: $p_\varphi(s_T,g)$ estimates whether the achieved outcome $s_T$ is close to the original goal $g$. 
Second, it supplies the target for corrective supervision. 
When $p_\varphi(s_T,g)$ is low, the trajectory is treated as evidence that the sampled actions were poor choices for reaching $g$, even though the same trajectory may still be useful as a positive demonstration for a relabelled goal $g'$. 
Thus, negative feedback in GCSL-NF is not an external reward penalty. 
It is the supervised learning signal obtained by evaluating an experience with respect to the original goal, in addition to evaluating it positively with respect to hindsight relabelled goals.
}

\section{Goal-Conditioned Supervised Learning with Negative Feedback}

In this section, we start with an illustrative example to explain the motivation behind our approach. Then we formalize this idea as two coupled learning problems: learning the goal-conditioned action classifier $p_\theta(\textrm{succ}(g)=1|s,a)$ and learning the state similarity function $p_\varphi$.
This way GCSL-NF can perform a dual evaluation of each trajectory:
the same experience is used positively as a target with respect to relabelled future states and correctively through the learned similarity $p_\varphi(s_T,g)$ with respect to the original goal, thus resolving the limitation inherent in GCSL.

\subsection{A motivating example}
\begin{figure}[!htbp]
  \centering
  \includegraphics[width=4.7in]{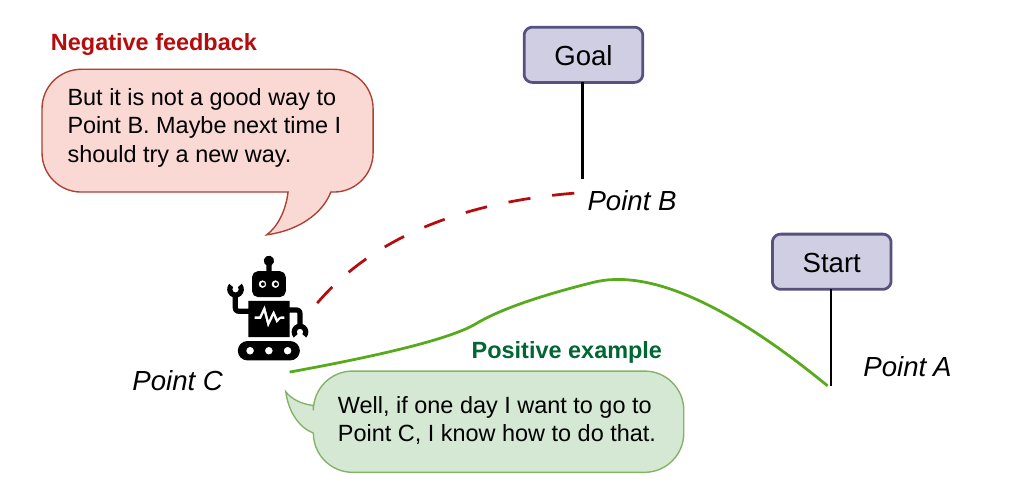}
  \caption{A Motivating Example: the agent starts from \textit{Point A},  initiates its journey with the objective of reaching \textit{point B}, but ultimately arrives at \textit{point C}. The green line indicates the actual path, while red dotted lines represent the distance between the achieved state and the original goal.}
  \label{fig1}
  \end{figure}

{To highlight the limitations of GCSL, consider the space robot from the introduction navigating between two sites, viewed as a two-dimensional abstraction of its terrain.}
The agent begins at some \textit{Point A} with the objective of reaching \textit{Point B}. However, due to the shortcomings of its current policy, the agent ends up at \textit{Point C} instead (see Figure~\ref{fig1}).

GCSL hinges on the notion that a trajectory failing to reach its intended goal effectively constitutes a successful path towards the state it actually attains. Thus, suboptimal trajectories can be reinterpreted as optimal by reassigning their goals. As illustrated in Figure~\ref{fig1}, this process involves redefining \textit{Point C} as the target, thereby transforming the trajectory represented by the green line into a form of expert guidance for achieving this new goal.

Crucially, this reinterpretation is not the sole lesson to be drawn here. In fact, there are two insights to be gained: while the green line denotes an optimal route to the newly assigned goal \textit{Point C}, it does not represent the most efficient path to the original goal \textit{Point B}. GCSL capitalizes on the former perspective to refine the policy but ignores the latter, which could provide valuable cues for future exploration. We turn GCSL into GCSL-NF by  incorporating this additional insight into the learning framework. Specifically, for each trajectory $\tau=\{s_1, a_1, s_2, a_2, \cdots, s_T, a_T\}$ generated under the current policy $\pi(a|s,g)$ and a specified goal $g$, we analyze it from the perspectives of both the relabelled goal $g'$ and the original goal $g$. For any given timestep $t$, if action $a_t$ facilitates reaching the relabelled goal $g'=s_{t+i}$, we enhance the likelihood of selecting $a_t$ at state $s_t$ with goal $g'$.
Concurrently, the suitability of $a_t$ in achieving the original goal $g$ is assessed based on the proximity of the final state $s_T$ to goal $g$, utilizing a similarity function. This dual approach allows the agent to either refine its policy or explore more.
The comparison of these different feedback types is illustrated in Section \ref{sec:feedback-ablation}.

\subsection{Proposed algorithm}
\begin{figure*}[!htbp]
    \centering
\includegraphics[width=0.9\textwidth]{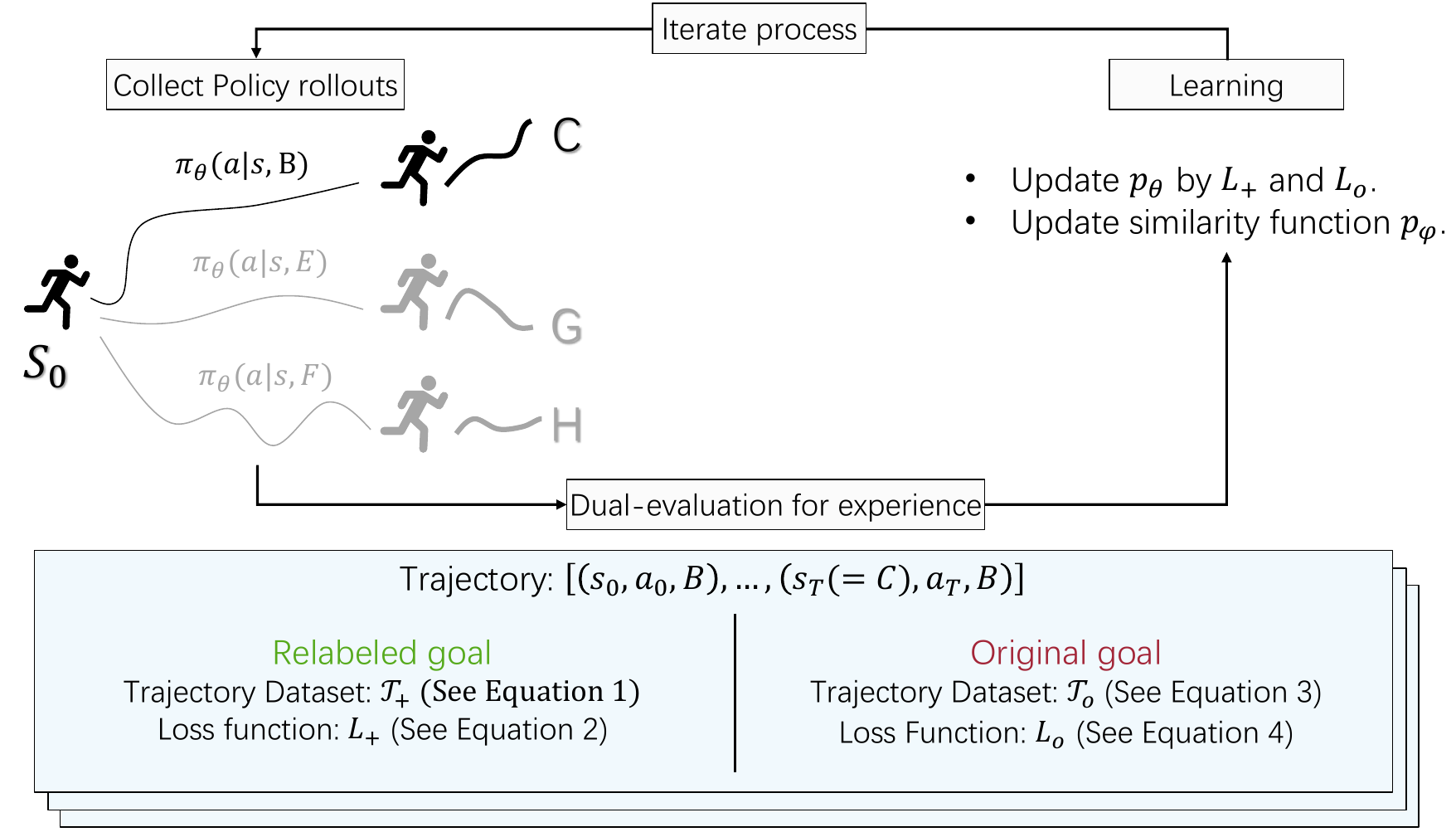}
\caption{Diagram of GCSL-NF: The agent learns how to reach goals by sampling trajectories, reconsidering the trajectories with both relabelled goals and original goals, and then updating the policy and similarity function.} \label{fig2}
\end{figure*}

{
Our approach relies on learning a parametrized function $p_\theta$ of actions $a_t$, states $s_t$, and goals $g$, which represents the success probability $p_\theta(\textrm{succ}(g)=1|s_t,a_t)$ of reaching a final state $s_T$ in the neighbourhood of the goal $g$ from state $s_t$ by acting according to $a_t$ (Step \ref{stepQ} below). In particular, we learn an approximate similarity function $p_\varphi$ between arbitrary states $s,s'$, which allows us to use $p_\varphi(s_T,g)$ as targets for $p_\theta(\textrm{succ}(g)=1|s_t,a_t)$. Figure~\ref{fig2} outlines our approach. The implementation is available on GitHub}\footnote{\url{https://github.com/zeqiang-zhang/gcsl_nf}}.

In the following, we list the steps in more detail:

\begin{enumerate}

\item Sample a goal $g$ from the goal space $\mathcal{G}$ using a fixed distribution $p$ on $\mathcal G$ {(Point B in Figure~\ref{fig1}).}

\item Generate a trajectory $\tau_g = (s_0, a_0, s_1, a_1, \cdots, s_T, a_T)$ from $p_\theta$ by choosing actions according to $$a_t =  \argmax_{a\in\mathcal A} p_\theta(\textrm{succ}(g)=1|s_t,a)$$ for $T$ steps. 

\item Add the trajectory $\tau_g$ to the replay buffer $\mathcal{R}$. 

\item Transform trajectories from $\mathcal{R}$ into expert tuples by relabelling, which are used as positive examples for imitation learning,
\begin{equation}
    \label{eq:T_+}
    \mathcal{T}_{+} \coloneqq \{(s_t, a_t, g' = s_{t+i}): t\geq 0,i>0, t+i \leqslant T\},
\end{equation}
with states $s_t$, corresponding actions $a_t$, and future states $s_{t+i}$ as relabelled goals $g'$ {(Point C in Figure~\ref{fig1})}. Notice, that not only the last state $s_T$ is considered as a goal, but also any state $s_{t+i}$ subsequent to the state $s_t$ is valid for hindsight relabelling.

\item \label{stepQ} Imitate the expert tuples through supervised learning by minimizing the loss function
    \begin{equation}
      \label{eq:L_+}
    L_+(\theta) = \mathbb{E}_{(s_t,a_t,g')\sim\mathcal{T}_{+}}\bigg[\underbrace{\vphantom{\sum_{a \in \mathcal{A}}}H(p_\theta(\textrm{succ}(g')=1|s_t,a_t),1)}_{\text{Imitation learning}} + \underbrace{ \alpha \sum_{a \in \mathcal{A}} H(p_\theta(\textrm{succ}(g')=1|s_t,a),0)}_{\text{Regularization}}\bigg],
    \end{equation}
where $H(x,y) = - y \log x - (1-y) \log (1-x)$ denotes the binary cross entropy between $x$ and $y$ (c.f. the discussion in Section \ref{RelatedWork}) and $\alpha$ is the regularization coefficient. 

\item Sample trajectories from the replay buffer $\mathcal{R}$ to create the dataset of tuples $\mathcal{T}_o$ with the original goals {(Point B in Figure~\ref{fig1})} according to
\begin{equation}
    \label{eq:T_o}
    \mathcal{T}_{o} = \{(s_t, a_t, s_T, g): t>0\},
\end{equation}
with states $s_t$, corresponding actions $a_t$, the final achieved state $s_T$, the original goals $g$. 

\item Evaluate the tuples $\mathcal{T}_{o}$ by the similarity function $p_{\varphi}(s_T,g)$  (detailed further in Section \ref{sec:distance}). 
\item Compute the loss $L_o$ over the tuples $\mathcal{T}_{o}$ with the estimated similarity:
\begin{equation}
    \label{eq:L_o}
    L_o(\theta) =  \mathbb{E}_{\mathcal{T}_{o}}\big[\gamma^{T-t} \, H\big(p_\theta(\textrm{succ}(g)=1|s_t,a_t), p_{\varphi}(s_T, g)\big)\big],
\end{equation}
where $\gamma^{T-t}$ exponentially discounts the loss according to the discount factor $\gamma$ and the remaining steps $T-t$.
 
\item Update the policy $\pi_{\theta}$ by minimizing the combined loss:
\begin{equation}
    \label{eq:update_pi}
    \theta_{new} \leftarrow \argmin_{\theta} \big(\beta_1 L_+(\theta) + \beta_2 L_o(\theta) \big),
\end{equation}
where $\beta_1$ and $\beta_2$ are the hyperparameters to balance the two losses. We set them to $1$. 

\item \label{stepP} Update the parameters $\varphi$ of the similarity function $p_{\varphi}$ according to Section \ref{sec:distance}.

\end{enumerate}

\subsection{Extension to continuous action spaces}
\label{sec:continuous-method}

{The formulation above assumes a discrete action space, in which the regularisation term of Equation~\eqref{eq:L_+} sums over all actions. We now extend the method to continuous action spaces following the deterministic policy gradient framework; the corresponding experiments are reported in Section~\ref{appendix:continuous}.}

In order to manage continuous action spaces, we utilize a similar approach as Deep Deterministic Policy Gradient (DDPG) \citep{silver2014deterministic,lillicrap2015continuous}. Like the discrete case, the approach for continuous action spaces also includes a parametrized function $p_{\theta_1}$ of action, state, and goal. The function $p_{\theta_1}$ is trained to imitate the expert tuples by minimizing the loss function
\begin{equation}
      \label{eq:continuous_L_+}
    L_+(\theta_1) = \mathbb{E}_{(s_t,a_t,g')\sim\mathcal{T}_{+}}\bigg[\underbrace{H(p_{\theta_1}(\textrm{succ}(g')=1|s_t,a_t),1 )}_{\text{Imitation learning}} + \underbrace{ \alpha H(p_{\theta_1}(\textrm{succ}(g')=1|s_t,\pi_{\theta_2}(s_t, g')),0)}_{\text{Regularization}}\bigg],
\end{equation}
 where $\alpha$ denotes the regularization coefficient, and $H(x,y)$ denotes the binary cross entropy. Note that, while in the discrete case the regularization term consists of the sum over all possible actions \eqref{eq:L_+}, here only the action generated by the policy $\pi_{\theta_2}$ is used. The loss $L_o$ is defined as the same as in the discrete case, and the function $p_{\theta_1}$ is updated by minimizing the combined loss $\beta_1 L_+(\theta_1) + \beta_2 L_o(\theta_1)$.
 
 The parameterized function $p_{\theta_1}(\textrm{succ}(g)=1|s_t,a_t)$ is used to train the deterministic policy function $\pi_{\theta_2}(s_t, g)$. The loss for the actor is defined by evaluating the proposed actions with the function $p_{\theta_1}$:
 \begin{equation}
    \label{eq:continuous_L_a}
    L_a(\theta_2) =  \mathbb{E}_{(s,s')\sim\mathcal{R}}\big[-p_{\theta_1}(\textrm{succ}(s')=1|s,\pi_{\theta_2}(s,s'))\big].
\end{equation}

 The similarity function $p_{\varphi}$ is trained in the same way as in the discrete case, according to Section \ref{sec:distance}.

\subsection{Similarity function \texorpdfstring{$p_\varphi$}{p\textunderscore phi}}
\label{sec:distance}
\begin{figure}[!htbp]
    \centering
\includegraphics[width=3.5in]{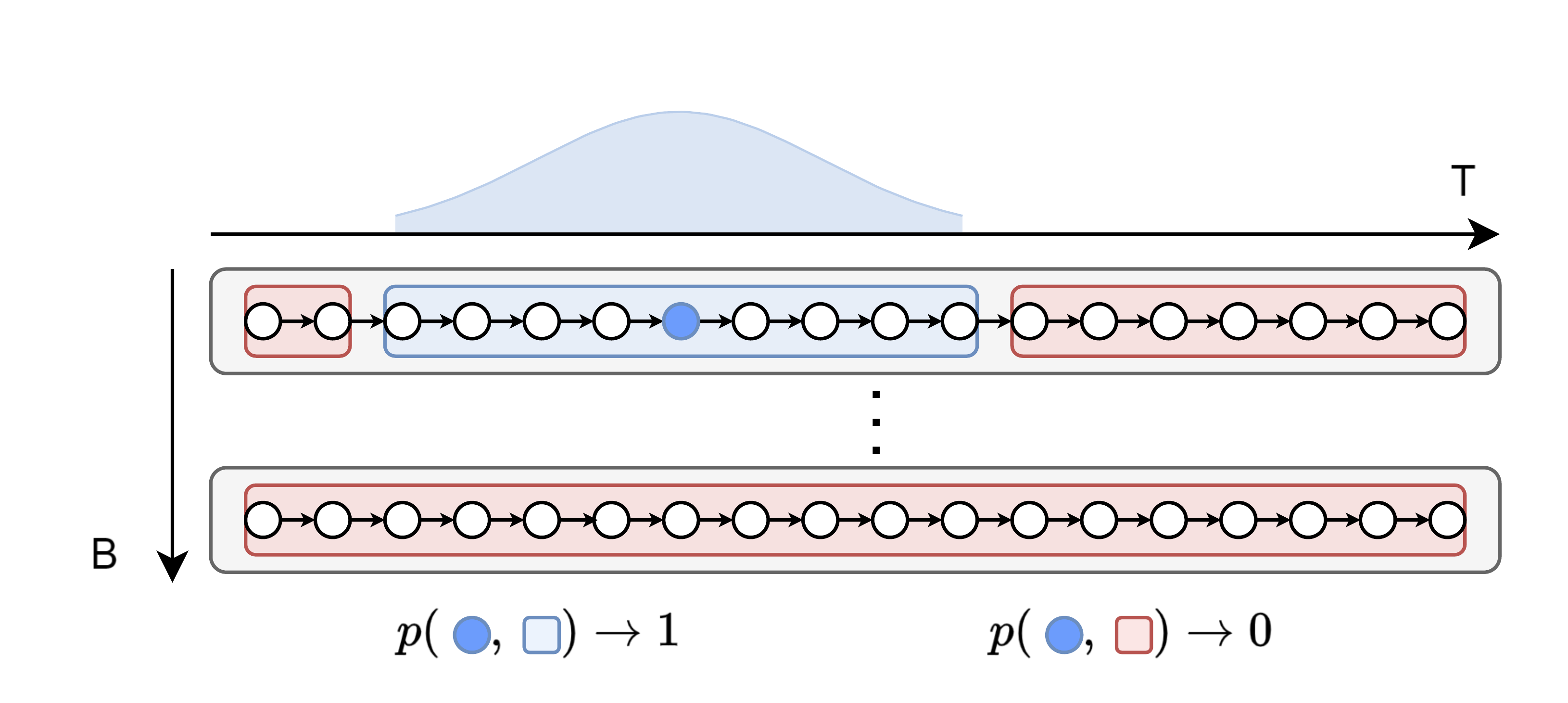}
\caption{Illustration of the learning process for $p_\varphi$ from trajectories. The figure shows how the neighbourhood relationship is defined on trajectories. The vertical axis indicates different trajectories, and the horizontal axis indicates time steps within the same trajectory. The blue region corresponds to positive samples from a unimodal distribution around the reference state, and the red region corresponds to negative samples.}
\label{fig3}
\end{figure}

In this Section, we introduce a method of learning similarity functions between states, using Monte Carlo methods  \citep{metropolis1949monte} and contrastive learning. Broadly speaking, $p_\varphi(s, s')$ is trained to approximate the probability that the distance between two arbitrary states $s,s'$ is smaller than a given threshold (see Figure~\ref{fig3}).

Given the unknown structure and transition dynamic of the environment beforehand, it is impractical to sample states based on their actual distances. Therefore, we devise an approximation strategy which involves sampling states from collected trajectories. State pairs are sampled under three distinct types of constraints: 
\begin{enumerate}
\item Positive pairs in $\mathcal{D}_+$ are extracted from the same trajectory, utilizing a unimodal distribution centered around the reference state:
\begin{equation}
    \label{eq:D+}
    \mathcal{D}_+ = \{(s_{i}, s_{j}): s_i, s_j \in \tau, \tau \in \mathcal{R}, j=\text{round}(j'), j' \sim \Delta (i,n)\},
\end{equation}
where $i$ and $j$ represent time steps, and $\Delta (i,n)$ denotes a symmetric triangle distribution with mode $i$ and threshold $n$. The triangular distribution is selected for its inherent boundary at the edge.

\item Negative pairs in $\mathcal{D}_-^1$, drawn from the same trajectory, are defined as:
\begin{equation}
    \label{eq:D-1}
    \mathcal{D}_-^1 = \{(s_{i}, s_{j}):s_i, s_j \in \tau, \tau \in \mathcal{R}, |i-j|>n\},
\end{equation}
where $i$ and $j$ represent time steps. To qualify as negative pairs, the states must be significantly separated from each other, achieved by imposing a threshold $n$ that specifies the minimum step distance between them. 

\item Negative pairs in $\mathcal{D}_-^2$, selected from distinct trajectories, are defined as:
\begin{equation}
    \label{eq:D-2}
    \mathcal{D}_-^2 = \{(s_{i}, s_{j}): s_i \in \tau, s_j \in \tilde{\tau}, \tau \neq \tilde{\tau} \in \mathcal{R}\},
\end{equation}
where $\tau$ and $\tilde{\tau}$ represent different trajectories, and $i$ and $j$ denote the respective time steps.

\end{enumerate}

The function $p_{\varphi}$ is trained to represent the probability of a given state pair to be close by minimizing the NCE loss  \citep{gutmann10,ma2018noise,eysenbach2022contrastive} 
 \begin{equation}
    \label{eq:varphi}
    L(\varphi) = -\Big[\mathbb{E}_{(s,s')\sim\mathcal{D}_+}[\log p_{\varphi}(s,s')] \\+ \sum_{i=1,2}\mathbb{E}_{(s,s')\sim\mathcal{D}_-^i}[\log(1- p_{\varphi}(s,s'))]\Big].
 \end{equation}

{

\section{Experiments}
\label{sec:experiments}

We evaluate GCSL-NF in three big blocks (performance, mechanism analysis and robustness) split into nine research questions (RQ1--RQ9), which are
summarised together with the corresponding experiments  in
Table~\ref{tab:rqmap}. Section~\ref{sec:environments}
 first describes the test environments and baseline methods. Section~\ref{sec:performance} establishes goal-reaching
\emph{performance} against baselines, in discrete (RQ1) and continuous (RQ2)
action spaces. Section~\ref{sec:mechanism} analyses the \emph{mechanism} behind
this performance: an ablation separates the contributions of the two feedback
pathways (RQ3), and a component study examines the learned similarity function
that makes the negative pathway possible, comparing it with alternative learned
distances and testing its sensitivity to the generating policy and the
temporal neighbourhood (RQ4/5). Section~\ref{sec:robustness} tests the robustness under four different perturbations and constraints
(\emph{robustness}, RQ6--RQ9).  Each experimental subsection states its research question.

\begin{table}[t]
  \caption{Research questions and corresponding experiments.}
  \label{tab:rqmap}
  \begin{tabular}{@{}c p{0.72\linewidth} c@{}}
    \toprule
        & Research question & Evaluated in \\
    \midrule
    \multicolumn{3}{@{}l}{\emph{Goal-reaching performance}} \\[2pt]
    RQ1 & Does GCSL-NF match or surpass HER- and GCSL-based baselines on
          discrete-action goal-reaching tasks, especially when observation-space
          distance is a poor proxy for physical or temporal proximity (obstacles,
          bottlenecks, LiDAR)? & Sec.~\ref{sec:main-results} \\[2pt]
    RQ2 & Does GCSL-NF extend effectively to continuous action spaces? & Sec.~\ref{appendix:continuous} \\
    \midrule
    \multicolumn{3}{@{}l}{\emph{Mechanism analysis}} \\[2pt]
    RQ3 & Is the improvement attributable to \emph{combining} positive hindsight
          supervision with original-goal negative feedback, rather than to either
          signal alone? & Sec.~\ref{sec:feedback-ablation} \\[2pt]
   RQ4 & Does the trajectory-induced similarity $p_\varphi$ provide more useful
          corrective targets than alternative learned distance measures? & Sec.~\ref{sec:distance-choice} \\[2pt]
    RQ5 & Is the learned similarity stable across generating policies and the
          threshold $n$? & Sec.~\ref{sec:distance-sensitivity} \\
    \midrule
    \multicolumn{3}{@{}l}{\emph{Robustness and constraint learning}} \\[2pt]
    RQ6 &  Can GCSL-NF recover from a corrupted initial policy and re-adapt when the
      goal distribution changes at runtime? & Sec.~\ref{sec:robustness-bias} \\[2pt]
    RQ7 & Does the learning signal remain informative under sensor and actuation
          noise? & Sec.~\ref{sec:robustness-noise} \\[2pt]
    RQ8 & Can negative feedback support a local, state-dependent safety
          constraint that persists after training-time enforcement is removed? & Sec.~\ref{sec:robustness-constraint} \\[2pt]
    RQ9 & Can negative feedback enforce a history-dependent, trajectory-level
          mission rule? & Sec.~\ref{sec:history-dependent-mission-constraints} \\
    \bottomrule
  \end{tabular}
\end{table}
}

\subsection{Environments and baselines}
\label{sec:environments}
\label{sec:baselines}

\begin{figure}[htbp!]
    \centering
\includegraphics[width=4.5in]{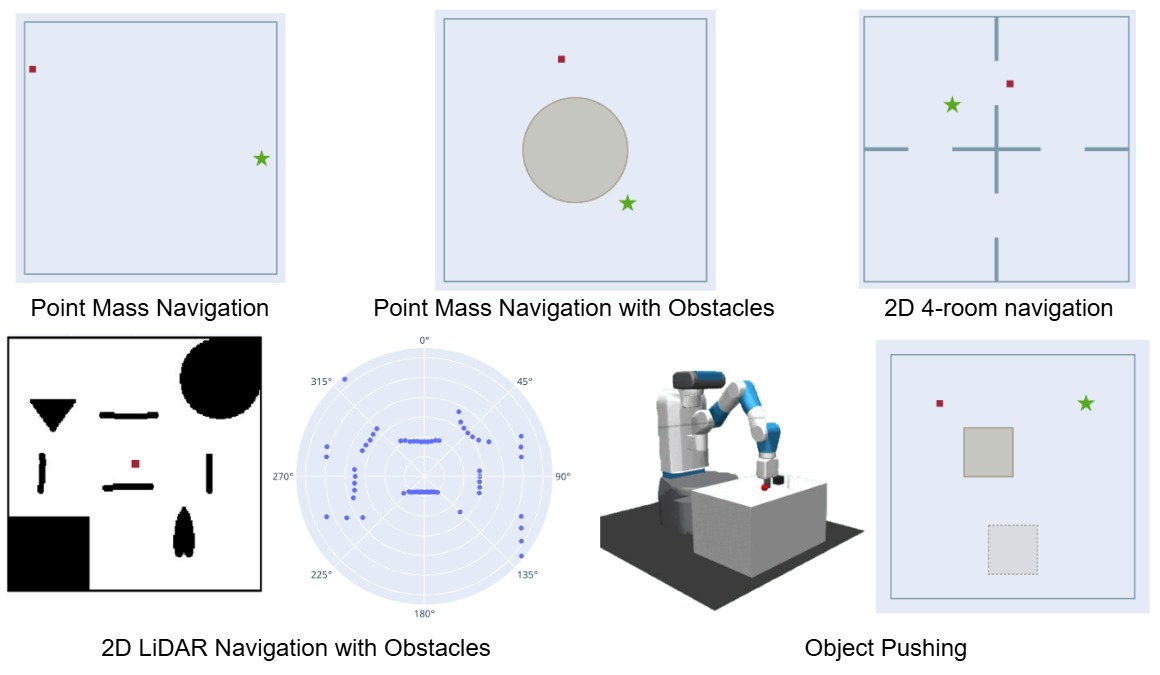}
\caption{Environments: red points indicate the agent position, and green stars denote the goal position.}
\label{fig4}
\end{figure}

We evaluate our method on five 2D goal-reaching or manipulation environments (Figure~\ref{fig4}), whose parameters are summarised in Table~\ref{tab:env-details} (Appendix~\ref{appendix:env-details}).
\emph{Point Mass Navigation} is the basic task of moving to a goal location.
\emph{Point Mass Navigation with Obstacles} adds a central circular obstacle that blocks movement on contact, distorting the optimal path.
\emph{2D 4-room Navigation} places the agent in four interconnected rooms where walls block movement on contact.
\emph{2D LiDAR Navigation with Obstacles} replaces coordinates with a $360^\circ$ LiDAR reading, so observation-space distance no longer reflects physical distance; goals are LiDAR observations at sampled poses $(x,y,\theta)$, and actions move or rotate the agent.
\emph{Object Pushing} is a customized discrete version of Fetch Push~\citep{plappert2018multi}: a robot arm must first reach a puck, push it to the goal, and then move to its own target, so the observation and goal cover both end-effector and puck coordinates.
Its continuous version is discussed in Section~\ref{appendix:continuous}.
{Viewed through the running scenario of Section~\ref{sec:introduction}, the suite covers the capabilities such a system needs: basic traversal, obstacle avoidance, perception through range sensing, and object manipulation.}

We compare against five baselines spanning HER-based reinforcement learning and GCSL-based supervised learning: HER DQN~\citep{andrychowicz2017hindsight}, a value-based method with hindsight experience replay; HER A2C~\citep{andrychowicz2017hindsight,mnih2016asynchronous}, its actor-critic counterpart; GCSL~\citep{ghosh2021learning}, self-imitation with hindsight relabelling; Weighted GCSL~\citep{yang2022rethinking}, which weights relabelled trajectories; and Contrastive GCRL~\citep{eysenbach2022contrastive}, which learns representations relating state-action pairs to future states. HER DQN and HER A2C require an external reward, which we define from the Euclidean distance between the achieved state and the goal. Algorithm details are in Appendix~\ref{comparison_algorithm}.

{

\subsection{Goal-reaching performance}
\label{sec:performance}

We first establish that GCSL-NF is competitive under standard training and testing conditions in both discrete and continuous action spaces.

\subsubsection{Discrete action spaces}
\label{sec:main-results}
\label{sec:discrete-main-comparison}

We evaluate GCSL-NF against the baselines in the five discrete-action environments of Section~\ref{sec:environments} (RQ1), paying particular attention to settings where observation-space distance is a poor proxy for physical or temporal proximity.

\begin{figure*}[!htbp]
    \centering
    \includegraphics[width=\textwidth]{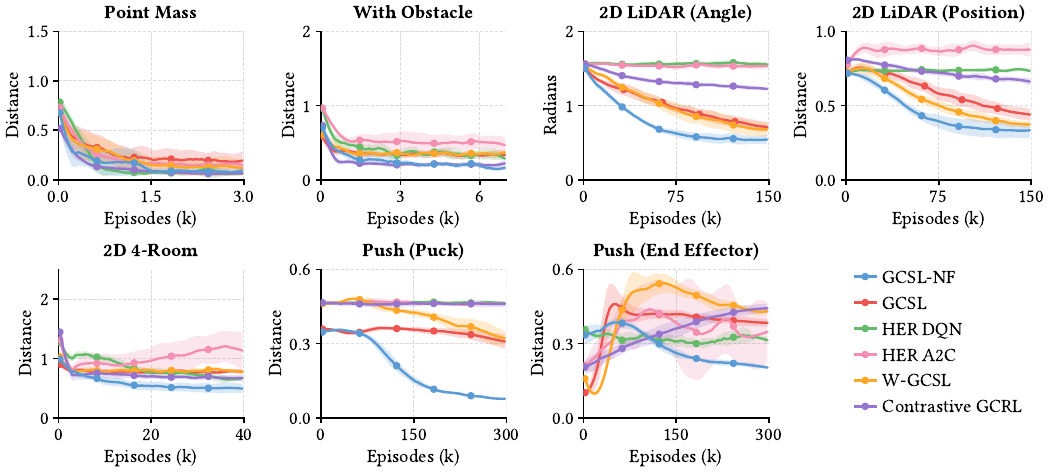}
    \caption{
    Discrete-action goal-reaching results across the benchmark environments.
    GCSL-NF matches or outperforms the baselines on the tested tasks.
    The results are averaged over $5$ random seeds, and the shaded area represents one standard deviation.
    }
    \label{fig:discrete-results}
\end{figure*}

Figure~\ref{fig:discrete-results} and Table~\ref{table:results} (Appendix~\ref{appendix_results}) show that GCSL-NF matches or surpasses the baselines in both learning efficiency and final performance.
On basic point-mass navigation all methods do well, since observation-space distance is well aligned with task progress.
The advantage grows with obstacles or non-trivial connectivity (point-mass navigation with obstacles, 2D 4-room navigation), where the learned similarity and original-goal feedback help the agent avoid local optima induced by misleading geometric proximity.
It is clearest in 2D LiDAR navigation: the observation is a LiDAR vector rather than a position, so the observation-space reward used by HER-based methods can mislead, whereas GCSL-NF learns its similarity from trajectory-induced neighbourhood relations and produces a more task-relevant signal.
In object pushing, GCSL-NF is the only method that reliably learns the manipulation behaviour, which requires an intermediate subgoal (reaching the puck before pushing it); negative feedback supplies corrective information when the achieved outcome remains far from the goal, helping the agent escape behaviours that merely move the end-effector. Section~\ref{sec:feedback-ablation} evaluates the two feedback pathways specifically.}

\subsubsection{Continuous action spaces}
\label{appendix:continuous}
\begin{figure}[t!]
    \centering
\includegraphics[width=\textwidth]{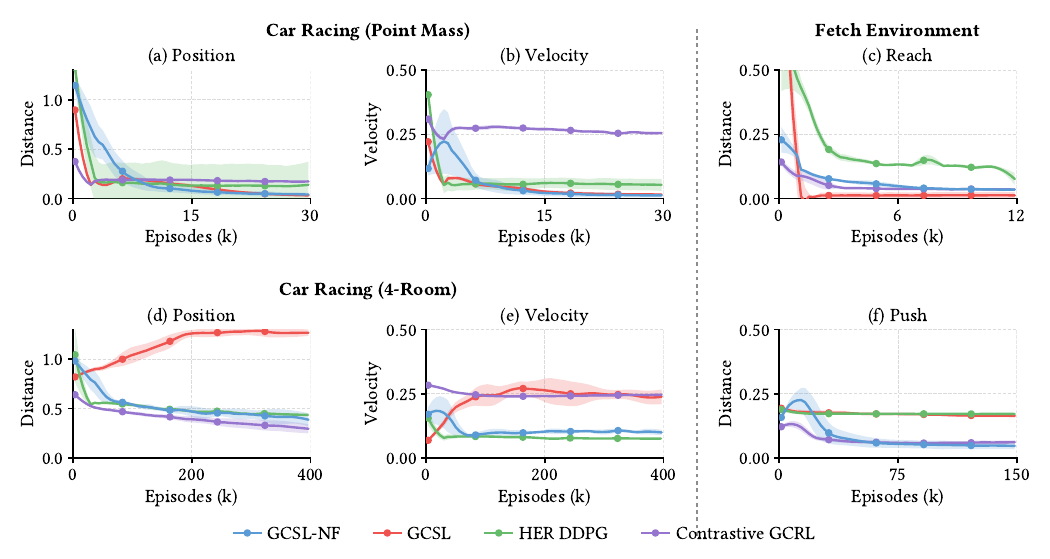}
\caption{Results: Comparison between continuous GCSL-NF, continuous GCSL, HER with DDPG and Contrastive GCRL. (a, b) Position and velocity distance in Car Racing (Point Mass); (d, e) the same in Car Racing (4-Room); (c) Fetch Reach; (f) Fetch Push. The results are averaged over 5 runs with different random seeds. The shaded area represents the standard deviation.}
\label{continuous}
\end{figure}

{We next address whether GCSL-NF extends effectively to continuous action spaces (RQ2), using the continuous formulation of Section~\ref{sec:continuous-method}.}
We consider two continuous-action variants of point mass navigation (without obstacles, and 4-room) and two standard manipulation tasks, Fetch Reach and Fetch Push~\citep{plappert2018multi, gymnasium_robotics2023github}.
In the point mass tasks the observation is four-dimensional (position and velocity in $x,y$) and the action controls acceleration rather than position directly, making them harder than their discrete counterparts; the goal specifies desired position and zero velocity.
For the Fetch tasks we use state-based observations with the default settings, where goals have different dimensions from the states.
We compare against continuous GCSL (extended from the discrete method~\citep{ghosh2021learning} by regressing on relabelled trajectories), Contrastive GCRL, and HER with DDPG~\citep{andrychowicz2017hindsight}.
Results are shown in Figure~\ref{continuous} and Table~\ref{table:results_continuous} (Appendix~\ref{appendix_results}).

All methods perform well on the car-racing point mass task (Figure~\ref{continuous}(a, b)). On the car-racing 4-room task (Figure~\ref{continuous}(d, e)), only HER DDPG, Contrastive GCRL, and GCSL-NF improve, while continuous GCSL collapses to a non-optimal policy for lack of exploration; DDPG explores through action noise, whereas GCSL-NF uses a deterministic policy in which exploration is implicitly driven by negative feedback. On Fetch Reach (Figure~\ref{continuous}(c)) all methods learn well (GCSL fastest), and on Fetch Push (Figure~\ref{continuous}(f)) GCSL-NF and Contrastive GCRL perform similarly while GCSL and HER DDPG fail to learn.

A notable difference is that Contrastive GCRL reaches the target position but tends to neglect local features that vary similarly across trajectories, such as velocity in the car-racing tasks (Figure~\ref{continuous}(b, e)). This follows from its negative-sampling scheme: because it contrasts future states against randomly sampled states from other trajectories, position dominates as the key distinguishing feature while local features are treated as noise, yielding an incomplete goal representation. Our method instead learns a local inverse distance function; although this can be slower, it ensures that all goal dimensions, not just position, are optimised.

{
\subsection{Mechanism analysis}
\label{sec:mechanism}

The previous section evaluated GCSL-NF at the policy level. We now analyse the mechanism behind these results.
Section~\ref{sec:feedback-ablation} isolates the contributions of the two feedback pathways (RQ3), and Sections~\ref{sec:distance-choice} and~\ref{sec:distance-sensitivity} examine the learned similarity function that makes the negative pathway possible, comparing it with alternative learned distance measures (RQ4) and testing its sensitivity to the generating policy and the threshold $n$ (RQ5).

\subsubsection{Ablation: positive and negative feedback}
\label{sec:feedback-ablation}

To isolate the two feedback pathways (RQ3), we compare three variants in the \textit{Point Mass Navigation with Obstacles} environment: (i) the full GCSL-NF objective, (ii) a positive-only variant using relabelled goals alone (standard hindsight imitation), and (iii) a negative-only variant using original-goal feedback alone.
All variants start without an initial collection of random trajectories, so the comparison isolates how each signal shapes learning from the agent's own rollouts (Figure~\ref{fig:feedback-ablation}).

\begin{figure*}[!htbp]
    \centering
    \includegraphics[width=\textwidth]{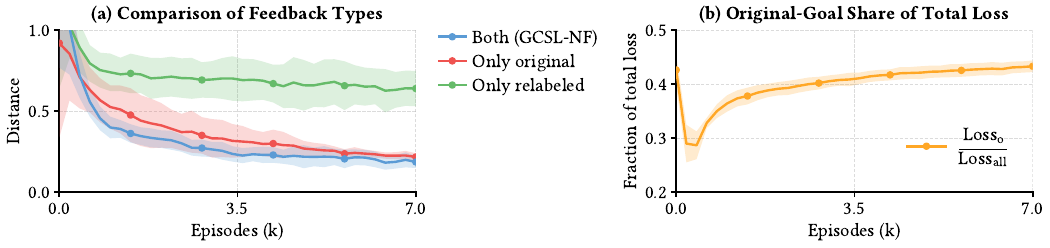}
    \caption{
    Ablation of positive and negative feedback in the \textit{Point Mass Navigation with Obstacles} environment.
    Panel (a) compares policies trained with positive feedback only, negative feedback only, and both feedback signals.
    Panel (b) shows the relative contribution of the original-goal loss, measured by $\frac{L_o}{L_o+L_+}$, over training.
    Results are averaged over $5$ random seeds, and the shaded area represents one standard deviation.
    }
    \label{fig:feedback-ablation}
\end{figure*}

The full objective outperforms both single-feedback variants (Figure~\ref{fig:feedback-ablation}(a)).
The positive-only variant gets stuck because failed attempts are always reinterpreted as successes for relabelled goals and carry no corrective information about the original goal, whereas the negative-only variant improves but converges slowly, lacking the dense positive supervision of hindsight relabelling.
The loss-ratio curve (Figure~\ref{fig:feedback-ablation}(b)) shows how the two interact: relabelled-goal supervision dominates early, when it provides dense positive examples, and the original-goal contribution grows as the policy improves and negative feedback becomes the main source of correction.
This supports the central design of GCSL-NF: positive supervision supplies stable imitation targets, while original-goal negative feedback prevents the policy from merely reinforcing what it already achieves.

\subsubsection{Comparison with alternative similarity measures}
\label{sec:distance-choice}
\label{sec:distance-analysis}

\begin{figure}[htbp!]
    \centering
\includegraphics[width=\textwidth]{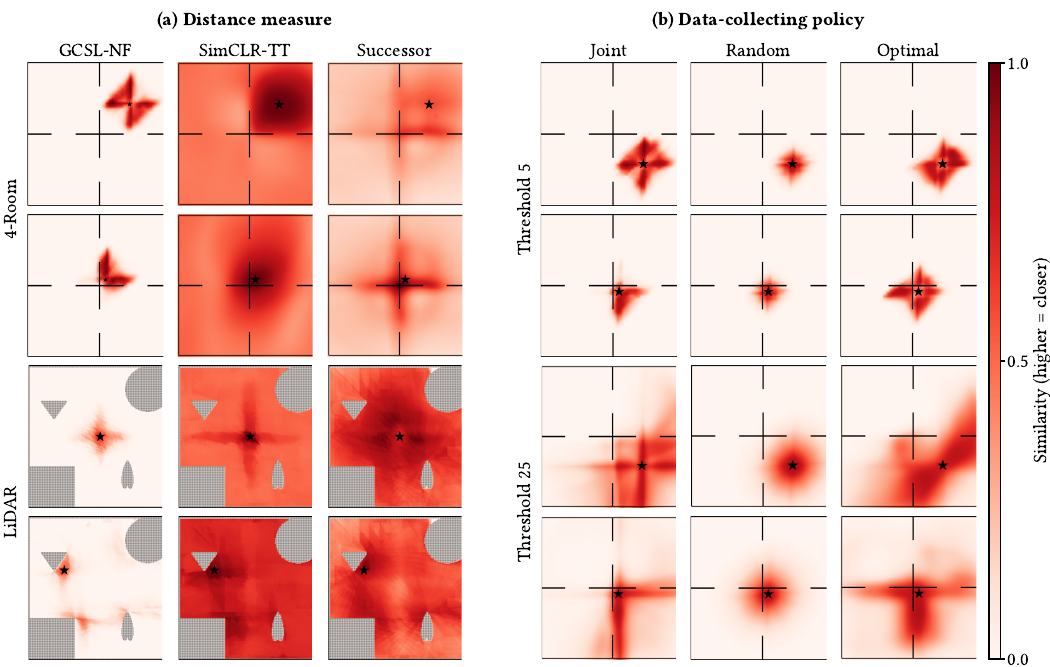}
\caption{Heatmap: (a) comparison between different kinds of distance measures,
and (b) comparison between inverse distance functions learned from different
data-collecting policies in the 4-room environment. Each panel shows the values
of an inverse distance function with respect to a fixed state, which is indicated by a
black star. In (a), the three columns show the inverse distance function proposed by
this paper, the distance function learned by SimCLR-TT, and the distance
function learned by the successor representation; the first two rows are the
results for the 4-room environment, and the last two rows are the results for
the LiDAR environment. In (b), the three columns show the inverse distance function
learned together with the policy, the inverse distance function learned by trajectories
generated by a random policy, and the inverse distance function learned by trajectories
generated by an optimal policy; the first two rows are the results for distance
functions with threshold $n=5$, and the last two rows are for threshold $n=25$.
Note that darker colour indicates closer proximity, because we have inverse
distance measures.}
\label{distanceheatmap}
\end{figure}

To understand the role of the learned similarity function (RQ4), we compare GCSL-NF against two alternative learned measures while keeping the rest of the algorithm fixed: a similarity learned by SimCLR-TT~\citep{schneider2021contrastive,chen2020simple}, and a successor-representation-based similarity~\citep{kulkarni2016deep,stachenfeld2017hippocampus}.
{SimCLR-TT treats temporally neighbouring states as positive pairs and measures cosine similarity between latent representations; the successor representation predicts expected discounted future state occupancy, normalised into a similarity score. Formal definitions are given in Appendix~\ref{appendix:similarity-defs} (Table~\ref{tab:table1}).}

\begin{figure}[htbp!]
    \centering
\includegraphics[width=\textwidth]{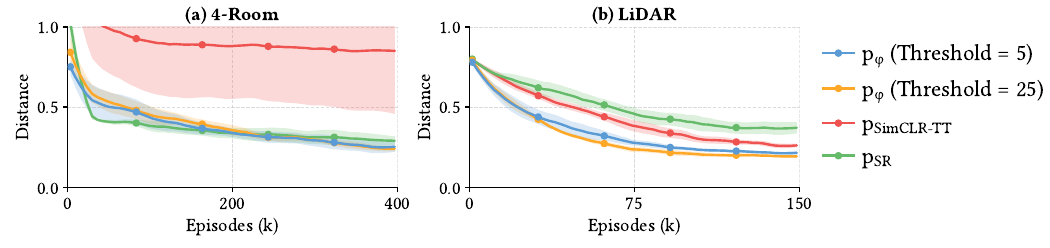}
\caption{{Comparison of learned distance measures and sensitivity to the threshold $n$, in the 2D 4-Room (a) and 2D LiDAR (b) navigation environments. 
GCSL-NF is trained with the trajectory-induced similarity $p_\varphi$ at two thresholds ($n=5$, used throughout the main experiments, and $n=25$), with a similarity learned by SimCLR-TT ($p_{\text{SimCLR-TT}}$), and with a successor-representation-based similarity ($p_{SR}$), keeping the rest of the algorithm fixed. 
The two thresholds are essentially indistinguishable in both environments.}}
\label{distance_performance}
\end{figure}

We evaluate the three measures in the 4-room and LiDAR environments, with distance maps in Figure~\ref{distanceheatmap}(a) and downstream results in Figure~\ref{distance_performance}(a) (4-room) and (b) (LiDAR).
Our similarity captures local spatial structure, while the alternatives are more global.
In the 4-room environment it learns the walls that are crucial for navigation; SimCLR-TT also locates the walls but still responds to distant states, and the successor representation captures the room structure but degrades near walls.
The gap is larger in the LiDAR environment, where the $L2$ distance in observation space no longer corresponds to distance in state space: our function still captures local structure, whereas SimCLR-TT and the successor representation become less effective locally, indicating that our measure is more robust when the observation space does not directly reflect the state space.
One shortcoming, shared by all three measures, is that similar observations at unrelated states can be overgeneralised, losing fine-grained structure. This is most visible in the LiDAR environment and could be mitigated by using memory or observation traces as input.

\subsubsection{Sensitivity to policy distribution and threshold \texorpdfstring{$n$}{n}}
\label{sec:distance-sensitivity}

Since $p_\varphi$ is learned from trajectories, it may in principle depend on the generating policy and on the temporal-neighbourhood threshold $n$.
We study both in the 2D 4-Room Navigation environment, where walls and bottlenecks make the distinction between Euclidean proximity and temporal reachability especially important.

The neighbourhood structure does depend on the generating policy: a random policy yields a similarity covering a smaller region than an optimal one. As the bottom panels of Figure~\ref{distanceheatmap}(b) show, however, this dependence is only pronounced when the temporal neighbourhood is defined very broadly and the policy is taken to an extreme (fully random vs.\ optimal); with the small thresholds used throughout the paper (top panels) the differences are negligible.
Quantitatively, Figure~\ref{distance_performance}(a, b) shows the learning curves for $n=5$ and $n=25$ to be essentially indistinguishable, so final performance is insensitive to $n$ within this range.
Together, these results support using a local trajectory-induced similarity as a stable corrective signal for negative feedback.
{
\subsection{Robustness and constraint learning}
\label{sec:robustness}

The experiments in Sections~\ref{sec:performance} and~\ref{sec:mechanism} evaluate goal-reaching performance and its mechanism under standard conditions. 
Deployed agents, however, face perturbations that are usually absent from standard benchmarks. 
A rover of the kind sketched in the introduction will encounter all four studied here: it starts from imperfect pre-mission training and is redirected to new objectives once deployed (RQ6, Section~\ref{sec:robustness-bias}), its sensors and actuators degrade in the field (RQ7, Section~\ref{sec:robustness-noise}), safety tethers available during commissioning are absent after deployment (RQ8, Section~\ref{sec:robustness-constraint}), and mission rules constrain admissible routes at the trajectory level (RQ9, Section~\ref{sec:history-dependent-mission-constraints}). 
In each case we test whether the learning mechanism remains informative once successful hindsight relabelling alone becomes unreliable, using both trajectory-relative contrastive structure and explicit negative feedback from failed, unsafe, or inadmissible behaviour.

\subsubsection{Recovery from policy bias and runtime goal distribution shift}
\label{sec:robustness-bias}
Adaptive systems can neither be assumed to begin from a neutral state nor to operate under fixed conditions: a deployed agent may inherit a corrupted prior, and the goals it is asked to achieve may change after deployment. 
We evaluate both cases.

For the first case, we use the Point Mass Navigation environment of Section~\ref{sec:environments}, updating the policy for $10$ gradient steps on a fixed ``right'' action label before training, after which the agent must reach goals sampled uniformly across the state space. 
The methods separate into three groups (Figure~\ref{fig:bias-recovery}(a)): GCSL-NF and Contrastive GCRL recover, almost reaching unbiased performance; HER A2C partially recovers with high seed variance; and GCSL, W-GCSL, and HER DQN fail to escape the bias. 
The entropy curves (Figure~\ref{fig:bias-recovery}(b)) explain why: the failing methods collapse to near-zero entropy almost immediately, whereas GCSL-NF and Contrastive GCRL show an initial spike, an exploration phase that breaks the bias before entropy decays again.

For the second case, we use the 2D 4-Room Navigation environment, drawing goals from a single room for $20$k episodes and then switching them without notification to the diagonally opposite room, retaining the policy and the replay buffer. 
The new goals require traversing two bottlenecks that the previous goal distribution barely exercised, so this is a runtime change of objective rather than a restart. 
All methods degrade at the change point (Figure~\ref{fig:bias-recovery}(c)), but GCSL-NF attains the lowest distance both before and after it and returns to its pre-switch level. HER DQN and Contrastive GCRL re-adapt to a higher distance, while W-GCSL and HER A2C settle clearly above their own pre-switch performance and GCSL remains weakest throughout.

Both cases expose the same failure mode: relabelling certifies behaviour that is wrong for the requested goals as correct for the goals it happens to achieve. 
Biased rightward trajectories become ``successes for rightward goals'', and after the switch, trajectories still ending in the old room become successes for old-room goals; W-GCSL amplifies this through advantage weighting, while HER DQN's reward under a misaligned policy is dominated by failures. 
GCSL-NF instead applies the original-goal loss $L_o$ (Equation~\ref{eq:L_o}), assigning an explicit negative signal to any trajectory that fails to reach the requested goal, which breaks the self-reinforcement loop even without successful trajectories. 
Contrastive GCRL escapes through a related route, using states from other trajectories as negative examples; both decouple the learning signal from what the policy currently achieves.

\begin{figure}[t]
\centering
\includegraphics[width=1\linewidth]{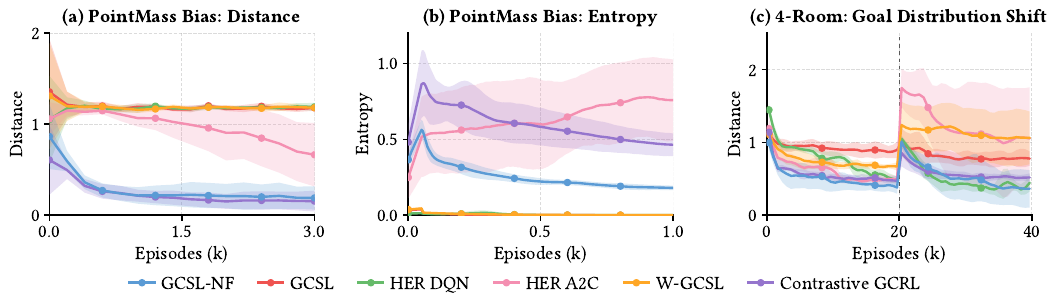}
\caption{Recovery from a corrupted prior and from a runtime goal change. 
(a) Distance to the goal and (b) action entropy under initial policy bias in the Point Mass Navigation environment, where the policy is pre-corrupted toward rightward movement before training.
(c) Re-adaptation in the 2D 4-Room Navigation environment, where the goal distribution shifts to the diagonally opposite room after $20$k episodes (dashed line), without resetting the policy or the replay buffer. 
Results are averaged over $5$ seeds; shaded areas represent one standard deviation.
}
\label{fig:bias-recovery}
\end{figure}

\subsubsection{Robustness to observation and actuation noise}
\label{sec:robustness-noise}

Real-world deployment involves sensor and actuation noise that is normally handled by filtering and state estimation across time. 

We use the 2D LiDAR Navigation environment for this sensitivity analysis and apply during both training and testing, (i)~\emph{sensor noise}, independent Gaussian noise $\mathcal{N}(0,\sigma^2)$ added to each of the $64$ LiDAR readings, and (ii)~\emph{actuation noise}, replacing the executed action by a uniformly random one with probability $p$. 
We sweep $\sigma, p \in \{2.5\%, 5\%, 7.5\%\}$ and compare against GCSL and W-GCSL, the only baselines that converge in the clean LiDAR setting (Section~\ref{sec:main-results}).

Figure~\ref{fig:lidar-noise} reports learning curves under sensor noise (a) and actuation noise (b) across all six conditions (5 seeds).
GCSL and W-GCSL improve marginally under mild sensor noise but fail as noise grows, whereas GCSL-NF retains meaningful learning that degrades gracefully, with a clear advantage in both position and orientation under mild to moderate sensor noise. 
Actuation noise is more damaging than sensor noise at matched levels, and at $7.5\%$ improvement becomes marginal for both noise types.

The baselines fail because they depend on absolute observation-space supervision. 
W-GCSL uses an advantage-weighted update with a sparse reward $r(s,g) = \mathbf{1}[\|s-g\|_2 < \tau]$; under observation noise, identical states are separated by an expected $\ell_2$ distance that grows with dimension and already exceeds $\tau$ at the mildest tested level, so the reward density collapses almost to zero,\footnote{For $\tau = 0.1$, $d = 64$, and $\sigma = 2.5\%$, the displacement of order $\sigma\sqrt{d}$ exceeds $\tau$, and the reward density drops from $\approx 0.64$ to $\approx 0.03$.} the value function loses its anchor, and the update reduces to noisy behavioural cloning. 
GCSL avoids this reward mechanism but has its relabelled imitation targets corrupted by the same noise, leading to near-uniform action probabilities and mode collapse at greedy deployment. 
GCSL-NF avoids both modes because its supervision comes from the contrastively learned $p_\varphi$ (Section~\ref{sec:distance}), which exploits relative neighbourhood structure within each trajectory rather than absolute distances; the signal stays informative as long as noise does not systematically reorder local temporal relationships.

\begin{figure}[t]
\centering
\includegraphics[width=\linewidth]{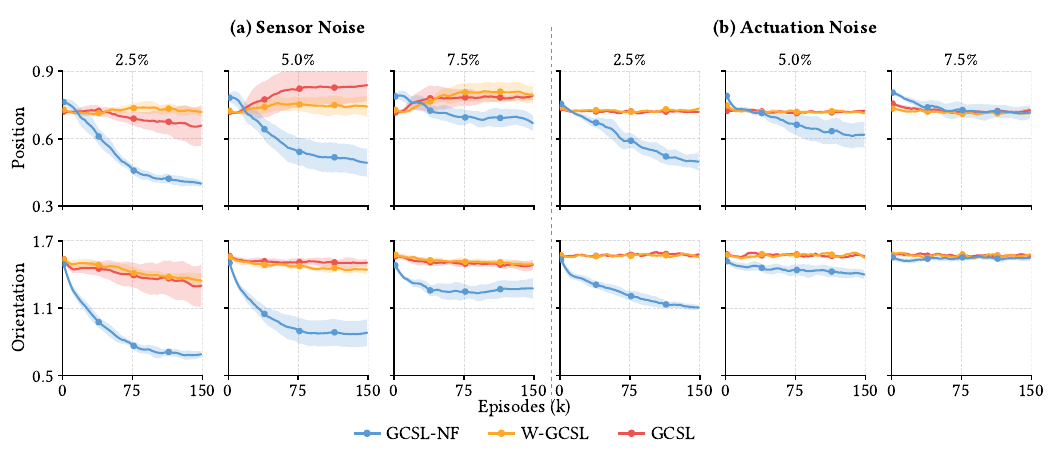}
\caption{Learning curves under (a) sensor noise and (b) actuation
noise in the 2D LiDAR Navigation environment, at three 
noise levels ($2.5\%$, $5\%$, $7.5\%$). Top row: position distance; 
bottom row: orientation distance. Results averaged over $5$ seeds; 
shaded area is one standard deviation.}
\label{fig:lidar-noise}
\end{figure}

\subsubsection{State-dependent safety constraints}
\label{sec:robustness-constraint}

\begin{figure}[htbp!]
    \centering
    \includegraphics[width=\textwidth]{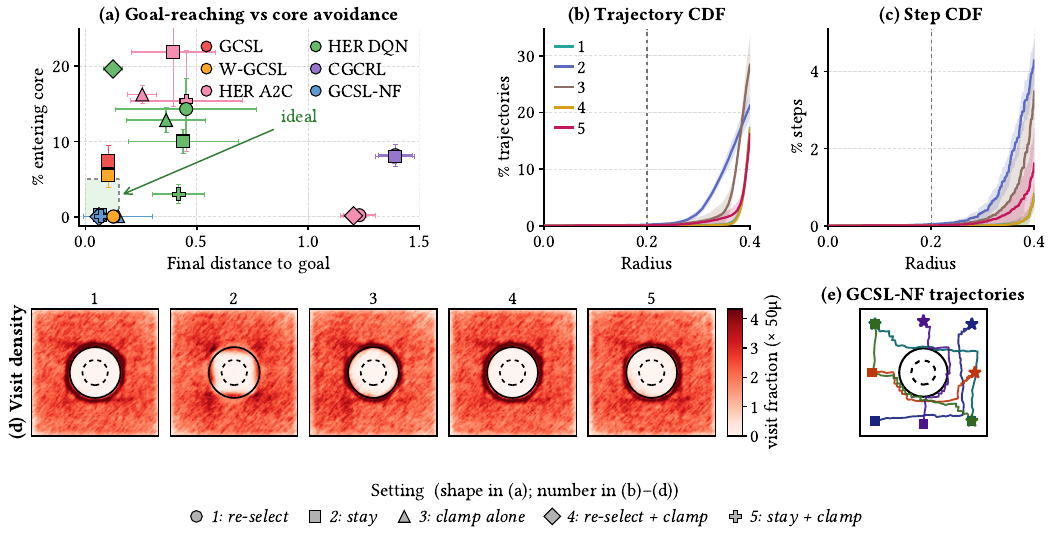}
    \caption{
    Danger-zone safety generalisation after training-time constraint exposure.
    Panel (a) compares goal-reaching accuracy and core-zone avoidance across six algorithms and five safety settings.
    For GCSL-NF, additional radial CDFs over trajectories (b) and steps (c), visit heatmaps (d), and representative trajectories (e) show how different combinations of action-space intervention and negative-feedback clamping affect unsafe-region avoidance. Evaluation is performed with all interventions disabled. Statistics are computed over $5{,}000$ test trajectories per method and setting, across $5$ random seeds.
    }
    \label{fig:danger-zone-safety}
\end{figure}

We next ask whether GCSL-NF can learn a safety constraint from training and preserve it once the constraint is no longer enforced at test time, motivated by settings where safety is imposed during training or supervised deployment but the deployed agent must later act without an external intervention mechanism.

We consider the Point Mass Navigation environment of Section~\ref{sec:environments} augmented with a circular danger zone of radius $0.4$ centred at the origin, with an inner core of radius $0.2$ marking the most hazardous region; start states and goals are sampled uniformly from outside the zone, but feasible paths between them often pass through it. 
During training, an action is treated as unsafe if its intended next state enters the hazardous region, following the safe-RL distinction between modifying the learning signal and modifying action selection~\citep{garcia2015comprehensive}. 
For GCSL-NF we fold this constraint into the learning signal by clamping the binary cross-entropy target of unsafe actions to zero, in both the relabelled positive branch and the original-goal negative branch, overriding the target from $p_\varphi$ whenever the transition would enter the zone. 
This does not alter the dynamics and adds no reward or penalty term. 
We combine this clamping with two training-time action-space interventions: \emph{re-select}, which executes the highest-valued safe alternative when the greedy action is unsafe (post-posed shielding~\citep{alshiekh2018safe}), and \emph{stay}, which blocks unsafe actions and leaves the agent in place (invalid-action masking~\citep{Huang2022closer}). 
This yields five GCSL-NF settings, \emph{re-select}, \emph{stay}, \emph{clamp alone}, \emph{re-select + clamp}, and \emph{stay + clamp}, which orthogonally vary whether safety is imposed through an action-space intervention, a learning signal, or both, connecting to intervention-based~\citep{saunders2018trial} and penalty-based~\citep{tessler2018reward} safe RL. 
All interventions are disabled at evaluation, so test-time behaviour measures whether the policy itself has learned the rule.

Figure~\ref{fig:danger-zone-safety}(a) reports the trade-off between goal reaching and danger-zone avoidance, using final distance to the goal and the fraction of trajectories entering the inner core; the lower-left region is desirable. 
The clamping settings apply only to GCSL-NF, since they require supervised binary targets that standard GCSL, W-GCSL, and Contrastive GCRL lack; those methods are evaluated under the action-intervention settings meaningful for them, and HER baselines can encode safety only through reward penalties. 
Almost all GCSL-NF variants reach the goal accurately while avoiding the core. 
GCSL and W-GCSL benefit from \emph{re-select} for core avoidance but stay less accurate than GCSL-NF and degrade under \emph{stay}; the HER baselines trade off inconsistently, either approaching the goal while still entering the hazard or avoiding the core only because they remain far from it; and Contrastive GCRL solves neither reliably. 
The radial CDFs (Figure~\ref{fig:danger-zone-safety}(b, c)), visit heatmaps (Figure~\ref{fig:danger-zone-safety}(d)), and representative trajectories (Figure~\ref{fig:danger-zone-safety}(e)) confirm that GCSL-NF routes around the zone even with all interventions disabled, indicating that the rule is learned by the policy rather than maintained by a runtime shield. 
This is not a formal safety guarantee, but it demonstrates a practically relevant form of adaptive safety under constraints that may change between training and deployment.

\subsubsection{History-dependent constraints}
\label{sec:history-dependent-mission-constraints}

We finally consider a constraint whose validity depends on trajectory history rather than a single state--action pair, testing whether negative feedback extends from local avoidance to simple trajectory-level rules. 
We introduce a no-obstacle task, \textsc{PointmassCorner} (Figure~\ref{fig:pointmass-corner-results}(a)): the domain $[-1,1]^2$ is a $4\times 4$ grid, and initial states and goals are sampled from the two opposite corner cells BL and TR, so every episode is a cross-corner traversal. 
A history-dependent waypoint rule makes the direct route inadmissible: BL$\rightarrow$TR trajectories must visit the top-left cell TL, and TR$\rightarrow$BL trajectories the bottom-right cell BR. 
Episodes last $100$ steps with per-step movement of magnitude $0.05$ and small Gaussian noise. All methods are trained for $150$k episodes with $2000$ initial exploration episodes and evaluated over $2000$ cross-corner trips per seed across $5$ random seeds.
Unlike the danger-zone constraint, which is checkable locally, correctness here depends on whether the required waypoint has already been visited: moving toward the goal is admissible after visiting it but inadmissible before, so the constraint is a trajectory-level mission rule rather than a state-level obstacle.

\begin{figure*}[t]
    \centering
    \includegraphics[width=\textwidth]{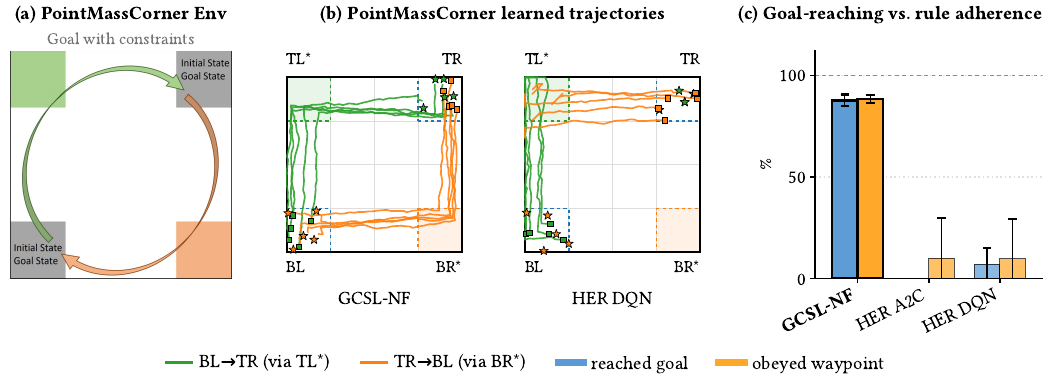}
    \caption{
    \textsc{PointmassCorner} task and learned behaviours under a history-dependent waypoint rule.
    (a) Environment schematic.
    Cross-corner trajectories must pass through a direction-dependent waypoint: TL for BL$\rightarrow$TR and BR for TR$\rightarrow$BL.
    (b) Representative trajectories of GCSL-NF and HER DQN under the trajectory-level constraint.
    Green trajectories correspond to BL$\rightarrow$TR trips via TL, and orange trajectories correspond to TR$\rightarrow$BL trips via BR.
    (c) Goal-reaching and waypoint-adherence rates.
    }
    \label{fig:pointmass-corner-results}
\end{figure*}

We incorporate the rule into GCSL-NF through the contrastive objective rather than a reward. 
\emph{Positive-relabel filtering} discards a relabelled sub-segment $s_i \rightarrow s_j$ if it represents a cross-corner transition that skips the required waypoint within $[i,j]$, preventing inadmissible shortcuts from being reinforced; \emph{whole-trajectory negative clamping} then clamps the target of a complete rule-violating trajectory to zero, treating it as explicit negative evidence. 
We compare against reward-based HER A2C and HER DQN, which cannot express the history-dependent rule as a per-step reward and so receive it as a trajectory-level delayed penalty subtracted from every transition of a violating trajectory in both original-goal and relabelled copies (applying it to relabelled copies is necessary, or HER would reinterpret a shortcut as a success for its own endpoint).

Figure~\ref{fig:pointmass-corner-results}(c) shows that GCSL-NF learns the constrained task more reliably than the reward baselines: its trajectories (Figure~\ref{fig:pointmass-corner-results}(b)) follow the TL-mediated route for BL$\rightarrow$TR and the BR-mediated route for TR$\rightarrow$BL, whereas the HER baselines face a difficult credit-assignment problem under the same budget and do not reliably solve it. 
The rule is encoded not as an obstacle or immediate reward but as a criterion determining whether relabelled experiences are reinforced or suppressed. 
In this experiment, the rule is given to GCSL-NF as a simple check that can be applied to a finished trajectory or to part of one. GCSL-NF uses this check only to decide which experiences to imitate and which to treat as negative examples. It does not turn formal temporal-logic rules into a policy, and it does not break a task into smaller subtasks or plan a sequence of goals. Handling full temporal-logic specifications and complex hierarchical planning is outside the scope of this work (Section~\ref{sec:limitations}).

Taken together, the four experiments point to one mechanism: negative feedback supplies corrective information that successful hindsight relabelling alone cannot, letting GCSL-NF learn from actions and trajectories that were unsuccessful, unsafe, or inadmissible with respect to the original goal or task rule. 
The results support a cautious conclusion: GCSL-NF offers no formal guarantees, but it lets a goal-conditioned supervised learner stay robust and keep respecting constraints when conditions change.
}

}

\section{Discussion}

\label{RelatedWork}
Our research addresses goal-conditioned tasks within the reinforcement learning (RL) domain, characterized by multi-goal settings and sparse rewards. { We first review related technical methods (Section~\ref{sec:rw-methods}) and then position our contribution within the broader autonomous and adaptive systems research community (Section~\ref{sec:positioning}).}

\subsection{Related Work in Goal-Conditioned Learning}
\label{sec:rw-methods}
Kaelbling \citep{kaelbling1993learning} first proposed the idea of hindsight relabelling, which augments the dataset with successful outcomes by reinterpreting failed trajectories. Hindsight relabelling approaches enhance data efficiency \citep{andrychowicz2017hindsight,rauber2017hindsight,ding2019goal,gupta2019relay} since trajectories generated by sub-optimal agents are still considered useful for training goal-conditioned policies, by simply pretending the actually reached states were the goals. This approach integrates well with other RL enhancements, such as curriculum learning \citep{fang2019curriculum}.

GCSL \citep{ghosh2021learning} adopts hindsight relabelling in conjunction with self-imitation learning, offering stability and compatibility with offline datasets. Different from self-imitation learning which usually chooses a subset of optimal trajectories to imitate \citep{10598576,oh2018self,hao2019independent} or learn a separate value function \citep{neumann2008fitted,abdolmaleki2018maximum,AWRPeng19}, GCSL maximizes data reuse by training on every collected trajectory. Advancements of GCSL aim to improve performance by assigning weights to relabelled trajectories \citep{yang2022rethinking,eysenbach2022imitating}. Drawing inspiration from these methods, our approach combines GCSL with negative feedback. Unlike HER-based methods that retain original trajectories, our strategy benefits from the precision and stability of supervised learning, obviating the need for explicit reward functions and thus broadening its applicability to autonomous learning across various tasks.

Our approach also utilizes similarity function learning within RL, where most strategies map state representations into a latent space and calculate the $l^2$-norm or cosine similarity between them, and then either directly predict distances or assess if the state distance falls below a certain threshold \citep{steccanella2022state,venkattaramanujam2019self,zang2022simsr}.
Value functions or Q-functions \citep{stachenfeld2017hippocampus,chane2021goal,nasiriany2019planning}, successor representations \citep{kulkarni2016deep,stachenfeld2017hippocampus}, and similarity-based approaches such as SimCLR-TT \citep{schneider2021contrastive,chen2020simple} can also be considered as kinds of distance functions in reinforcement learning. Diverging from these approaches, we employ Monte-Carlo based contrastive learning methods \citep{metropolis1949monte} to estimate state pair co-occurrence frequencies, a technique agnostic to observation structure, thereby ensuring wide applicability. Compared to these approaches, which primarily capture global structural relationships within the environment, our method excels at modeling local spatial structures. Since the agent can leverage hindsight relabeling to improve policy learning, providing feedback from a local inverse distance function can be more effective in guiding exploration; see Section \ref{sec:distance-analysis} for a detailed comparison.

In our work, we make use of contrastive learning, a method predominantly applied in unsupervised or self-supervised representation learning, where positive and negative examples are used to extract discriminative features from unlabeled data \citep{chen2020simple,chen2020big,he2020momentum,caron2020unsupervised}. Contrastive learning has shown success not only in computer vision but also in other fields, such as natural language processing \citep{mikolov2013distributed,chi2020infoxlm}, recommender systems \citep{yu2023self,huang2022self}, and reinforcement learning \citep{laskin2020curl, stooke2021decoupling, schneider2021contrastive, 9939161, eysenbach2022contrastive}. 
Instead of simply using contrastive learning to learn latent representations of states as a drop-in replacement for reconstruction or other representation learning methods, Contrastive Goal-Conditioned Reinforcement Learning (Contrastive GCRL) \citep{eysenbach2022contrastive} frames the goal-conditioned reinforcement learning problem \textit{as} contrastive learning, contrasting states reached within a trajectory against states from other trajectories to approximate a goal-conditioned Q-value function. {Although both methods build on contrastive learning, they differ fundamentally. Contrastive GCRL never uses the original goal: its negatives are undirected states drawn from other trajectories, and the contrast serves to shape a representation in which a Q-value is estimated. GCSL-NF instead contrasts the achieved outcome against the specific goal the agent intended but missed, and uses this directed signal to correct the policy directly. The original goal is therefore not merely a ``negative'' in our formulation: each trajectory is evaluated twice, positively with respect to relabelled goals and correctively with respect to the original goal.}

\begin{table*}[!htbp]
\caption{Comparison between different algorithms for goal-conditioned problems \label{tab:table_algos}}
\centering
\begin{tabular}{c|c|c|c|c|c}\hline

 & Vanilla RL & RL with HER & GCSL & Contrastive GCRL & GCSL-NF\\\hline
Relabelled goal       &                   & \checkmark            &    \checkmark          &  \checkmark            &     \checkmark                           \\\hline

Original goal    &  \checkmark &    \checkmark         &              &              &          \checkmark              \\  \hline    
Contrastive goal &       &      &              &           \checkmark   &              \checkmark                  \\\hline
Contrastive action         &       &      \checkmark       &  \checkmark            &              &              \checkmark                  \\\hline
External reward function &  \checkmark   &      \checkmark       &              &              &                     \\  \hline

\end{tabular}
\end{table*}

We compare and summarize the differences between widely used algorithms for goal-conditioned problems and our proposed method in Table \ref{tab:table_algos}. With the exception of standard goal-conditioned reinforcement learning, all listed algorithms incorporate goal relabeling to improve learning efficiency. However, only our method, vanilla RL and RL with HER utilize information from the original goal. Notably, our approach does not rely on an external reward function, distinguishing it from vanilla RL and HER. Furthermore, our method employs both random actions as negative examples during policy training (contrastive action) and random states as negative examples during similarity function training (contrastive goal). These mechanisms enhance learning stability and robustness, setting our approach apart from existing goal-conditioned methods.

{

\subsection{The broader picture:  GCSL-NF within self-adaptive systems}
\label{sec:positioning}

A central challenge in engineering self-adaptive systems (SAS) is design-time uncertainty: the precise effect of an adaptation action may not be known when the system is built \citep{Hezavehi2020, Calinescu2020uncertainty}. Online reinforcement learning has emerged as a way to address this, by employing RL at runtime so that the managing system learns suitable adaptations from actual operational data \citep{Gheibi2021,palm2020online,Metzger2022Realizing}. A fundamental challenge of this approach is that developers must explicitly specify a reward function, which might be straightforward in well-structured domains with predefined tasks, but difficult for unstructured open-ended environments, that are for example often faced by autonomous embodied agents outside purposefully built manufacturing halls. GCSL-NF addresses this scenario in that it requires no prespecified reward function. Supervision is derived in a self-supervised fashion from the agent's own trajectories — both those that reach their intended goal and, via the negative-feedback pathway, those that do not (Section~\ref{sec:distance}). Goals are expressed directly as target observations rather than as engineered reward signals. This removes the reward-specification step and replaces it with a learned, observation-relative similarity function. 

This reward-free property supports two complementary modes of use within a self-adaptive system, distinguished by where learning occurs relative to deployment.
In the pretraining mode, GCSL-NF produces goal-conditioned policies in a simulator at design time that are pretrained broadly. Because no objective is hard-coded, a single policy is triggered by a wide range of observational goals, and can be invoked by the managing system at runtime to realise a broad spectrum of goals determined by the analyse or plan stage \citep{kephart2003vision}. This way, GCSL-NF supplies a reusable adaptation capability that can be invoked by different goals at the execution stage, while goal selection remains with the managing system. 
In the online mode, the GCSL-NF learning loop is itself embedded in the managing system's feedback loop: the agent continually generates trajectories, observes successes and failures, relabels them, and refines its policy from operational experience. This is suited to settings where neither design-time reward specification nor runtime human intervention is feasible like our space robot example from the introduction. The same learning loop underlies both modes: what changes is not the algorithm but the conditions under which the loop runs, in particular whether the goals and the environment remain stationary. Sections~\ref{sec:performance} and~\ref{sec:mechanism} evaluate the loop under stationary conditions, and the goal-shift experiment in Section~\ref{sec:robustness-bias} evaluates re-adaptation when the objective changes at runtime. 

In the engineering disciplines, self-adaptive systems are usually studied with a strong focus on safety and stability, whereas the machine learning literature usually  emphasizes representation and learning. A common meeting ground is safe reinforcement 
learning, which addresses the problem of ensuring that learned policies 
satisfy safety constraints during exploration and 
deployment~\citep{garcia2015comprehensive}. Existing approaches include 
constrained Markov decision processes and Lagrangian-based policy 
optimisation~\citep{altman1999constrained, achiam2017constrained}, 
runtime shielding that overrides unsafe actions according to a formal 
specification~\citep{alshiekh2018safe}, control-barrier-function methods 
that exploit known dynamics~\citep{cheng2019end}, and recovery policies 
that intervene near unsafe regions~\citep{thananjeyan2021recovery}. 
Safe-RL methods and GCSL-NF operate at different layers of the autonomy stack: safe-RL machinery acts at the level of action selection or policy optimisation under externally specified constraints, while GCSL-NF acts at the level of the learning signal, deriving corrective information from the agent's own experience. This layering makes the two natural 
partners. A GCSL-NF learner can be placed beneath a runtime shield that 
guarantees no unsafe action is ever executed during deployment, with the 
shield providing the certifiable safety envelope and GCSL-NF supplying 
the underlying goal-reaching competence without per-goal reward 
engineering. Conversely, the negative-feedback mechanism can be combined 
with constrained policy optimisation by routing constraint violations 
into the negative-feedback objective itself, as we illustrate empirically 
in Section~\ref{sec:robustness-constraint}, where unsafe transitions are 
treated as supervised negative targets and the resulting policy continues 
to avoid the unsafe region even when external enforcement is removed. 
The same composition pattern extends to history-dependent mission 
constraints (Section~\ref{sec:history-dependent-mission-constraints}): rules that cannot be 
checked locally can be enforced during training through trajectory-level 
negative feedback and combined with a runtime monitor that handles the 
residual cases. In each instance, GCSL-NF reduces the reward-engineering 
burden on the safety-oriented layer without replacing the guarantees it 
provides.

\subsection{Limitations and threats to validity}
\label{sec:limitations}

It is important to note that GCSL-NF 
does not provide formal safety guarantees, certified satisfaction of 
arbitrary temporal-logic specifications, structural runtime 
reconfiguration, or end-to-end sim-to-real validation. The behaviour we demonstrate in Section~\ref{sec:robustness} is empirical properties of the learned 
policy rather than verified system-level claims. For safety-critical 
deployments, the method is therefore best understood as a component that supplies goal-reaching competence and empirical 
constraint-awareness to a larger architecture, with formal guarantees 
delivered by complementary components such as 
shielding~\citep{alshiekh2018safe}, constrained policy 
optimisation~\citep{achiam2017constrained}, barrier-function 
methods~\citep{cheng2019end}, or temporal-logic-based 
synthesis~\citep{hasanbeig2018logically}. Integrating GCSL-NF with abstract, 
multi-objective, or hierarchically structured goal specifications is a 
natural direction for follow-up work. Another further direction is to combine the self-supervised negative-feedback signal 
with domain-specific external rewards where such rewards are available, using 
the learned similarity for goal-reaching competence and the reward for 
additional task preferences.

GCSL-NF is currently formulated for goals that can be represented in the agent's observation or goal space, and its similarity function is learned from trajectory-induced neighbourhood relations. 
This makes the method directly applicable to state-based goal-reaching tasks and, as shown in Sections~\ref{sec:robustness-constraint}--\ref{sec:history-dependent-mission-constraints}, extensible to representative constraint-driven settings through mechanisms such as action-level clamping and positive-relabel filtering. 
However, these experiments should not be interpreted as solving general reasoning required for complex missions.
The current framework does not yet handle arbitrary temporal-logic specifications, hierarchical task decompositions, multi-objective trade-offs, or vague requirements that cannot be mapped to observable states, waypoints, or rule-checkable trajectory properties. 
Future extensions may address this by integrating concrete-to-abstract goal embeddings~\citep{Wurzberger2026concrete}, temporal-logic specifications~\citep{hasanbeig2018logically}, or hierarchical goal representations.

The current framework utilises a local similarity function $p_\varphi$ together with a uniform goal-sampling strategy. 
Although this provides useful corrective information, the negative-feedback signal does not by itself identify the causal source of failure. 
In open autonomous systems, a failed trajectory may result from inappropriate actions, external disturbances, partial observability, actuator faults, infeasible goals, or goal mis-specification. 
Distinguishing these cases would require additional mechanisms for uncertainty estimation, failure diagnosis, or causal attribution. 
Meanwhile, uniform goal sampling does not adapt to the agent's evolving competence, limiting sample efficiency. 
Randomly sampled goals may be too difficult early in training, too easy later in training, or poorly aligned with bottlenecks and frontier regions that are important for exploration. 
Recent work on temporal-distance-aware representations for unsupervised goal-conditioned RL illustrates how learned temporal distance can be used to select exploratory goals that are reachable but still informative, and ultimately improve goal-reaching behaviour~\citep{bae2024tldr}. 
Integrating active goal selection, automatic curricula, and memory-conditioned or uncertainty-aware distance metrics represents a promising direction for improving exploration and performance.

We close by considering four threats to validity. 
Regarding construct validity, goal achievement is measured through 
task-specific proxies such as final distance, orientation error, and 
constraint-violation rates, which capture the intended outcome but not whether a 
trajectory was efficient or safe throughout the episode, and which measure 
empirical adherence rather than formal verification. 
Regarding internal validity, although we use comparable architectures, 
training budgets, and evaluation protocols across methods, reward-based and 
supervised-relabelling paradigms respond differently to the same hyperparameter 
budget, so the comparison cannot fully exclude tuning effects; we mitigate this 
by using identical environment dynamics and evaluation goals throughout and by 
reporting learning curves with across-seed variability. 
Regarding conclusion validity, all results are averages over five random 
seeds, so our claims concern performance patterns that are consistent across 
seeds and tasks rather than the statistical significance of small numerical 
differences.

Furthermore, with respect to external validity, we note that all experiments were conducted in simulated environments, mostly involving low-dimensional navigation, LiDAR-style observations, or simplified continuous-control tasks. These environments were not chosen because the proposed method is limited to navigation. Goals can in principle be defined in any domain, including for example adaptive software systems where goals could be defined over system states like ``keep the response delay between x and y ms'' etc. Instead the environments were primarily designed to isolate the algorithmic role of negative feedback in autonomous embodied systems. However, it should be noted that our experiments do not capture the full complexity of deployed autonomous systems. In particular, the results should not be interpreted as evidence of direct transfer to high-dimensional robotic manipulation, image-based perception, real hardware, non-stationary dynamics, human interaction, or safety-critical cyber-physical systems. 
The robustness experiments in Section~\ref{sec:robustness} introduce biased initial policies, observation and actuation noise, training-time safety interventions, and history-dependent waypoint rules, but they remain controlled simulations. 
A full deployment validation would require sim-to-real transfer, hardware experiments, runtime monitoring, and formal safety analysis.
}

\section{Conclusion}
In this study, we introduced GCSL-NF, a learning scheme that overcomes a structural limitation of goal-conditioned supervised learning: by evaluating every trajectory both with respect to relabelled goals and with respect to the originally intended goal, the method turns failed attempts into corrective supervision instead of discarding them. 
The corrective targets are supplied by a trajectory-induced similarity function learned with Monte Carlo based contrastive estimation, so that no external reward function or geometric distance needs to be specified. 
Our empirical findings show that this dual evaluation reduces the impact of biases that approaches trained on self-generated data suffer from, matches or surpasses existing baselines across discrete and continuous goal-reaching tasks, and lets the agent re-adapt and keep respecting constraints when conditions change after training.
For autonomous systems that must learn in the absence of engineered reward functions, such as the space robot sketched in the introduction, every failed attempt then becomes supervision rather than discarded data. 
A promising future direction would be to develop sophisticated goal-sampling strategies, moving beyond the current reliance on randomly selected goals from the goal space to enable novelty-driven exploration and hence more autonomy in the learning process.

\section*{Acknowledgments}
Part of this study was funded by the European Research Council (ERC-StG-2015---ERC Starting Grant, Project ID:~678082, BRISC: Bounded Rationality in Sensorimotor Coordination).

\bibliography{bib}
\bibliographystyle{style}

\appendix
\section{Comparison of Algorithms}
\label{comparison_algorithm}

We provide comparisons to HER DQN, HER A2C, GCSL, Weighted GCSL, and Contrastive GCRL.

\begin{itemize}
    \item \textbf{HER DQN} \citep{andrychowicz2017hindsight} uses Q-learning to learn a policy that maximizes the expected return with hindsight experience replay \citep{andrychowicz2017hindsight}. The Q function is learned by minimizing the TD-error
    \begin{equation}
        \label{eq:TD-error}
        L_{Q}(\varphi) = \mathbb{E}_{\tau \sim \mathcal{D}}[(r_t + \gamma \max_{a'}Q_\varphi(s_{t+1}, a', g) - Q_\varphi(s_t, a_t, g))^2].
        \end{equation}
    We put the trajectories both with the relabelled goals $g'$ and the original goals $g$ into the replay buffer $\mathcal{D}$. The reward is defined based on the $L2$ distances between the current states and the goal $r_t = -\|s_t - g\|_2$. To stabilize the learning process, we use a target network to calculate the target value \citep{van2016deep}. The target network is updated by the main network after every trajectory. To encourage exploration, we use an $\epsilon$-greedy policy where $\epsilon$ is set to $0.001$. The discount factor $\gamma$ is set as $0.99$.

    \item \textbf{HER A2C} \citep{andrychowicz2017hindsight,mnih2016asynchronous} is an actor-critic method that optimizes a policy and a value function simultaneously \citep{mnih2016asynchronous}. The actor is trained to maximize the expected return by policy gradient with respect to an advantage function,
    \begin{equation}
        \label{eq:policy}
        L_{\pi}(\theta) = \mathbb{E}_{\tau \sim \mathcal{D}}[- \log \pi_{\theta}(a_t|s_t, g)\\\cdot \underbrace{(r_t+\gamma V_\varphi(s_{t+1},g)-V_\varphi(s_t,g))}_{\text{Advantage function}}],
    \end{equation}
    and the critic is trained to minimize the TD-error
    \begin{equation}
        L_{V}(\varphi) = \mathbb{E}_{\tau \sim \mathcal{D}}[(r_t + \gamma V_\varphi(s_{t+1},g) - V_\varphi(s_t,g))^2].
        \end{equation}
        We put the trajectories both with the relabelled goals $g'$ and the original goals $g$ into the replay buffer $\mathcal{D}$. The reward only depends on the Euclidean distances between the current states and the goal: $r_t = 1/(1+\|s_t - g\|_2)$. The discount factor $\gamma$ is set as $0.99$. To encourage exploration, we sample actions from a softmax-transformed policy.
    \item \textbf{GCSL} \citep{ghosh2021learning} considers the total experience $\mathcal{D}$ as successful trajectories by considering the actually achieved states as goals. When executing in all of the environments, the first $200$ trajectories are collected by executing a random policy to accumulate random trajectories, followed by a greedy policy that selects actions according to $\arg \max_a \pi (a|s,g)$.
    \item \textbf{Weighted GCSL} \citep{yang2022rethinking} is an advanced iteration of GCSL that optimizes performance by applying weights to relabelled trajectories. The loss function is
    \begin{equation}
        L_{\pi}(\theta)=\mathbb{E}_{\tau \sim \mathcal{D}}[-w_t \log \pi_{\theta}(a_t|s_t, g)],
    \end{equation}
    where $w_t$ is the weight of the trajectory defined as
    \begin{equation}
        w_t = \gamma^{h}\cdot\exp_{clip}(A(s_t,a_t,g))\cdot\epsilon (A(s_t,a_t,g)),
    \end{equation}
     where $A(s_t,a_t,g) = r_t+\gamma V_\varphi(s_{t+1},g)-V_\varphi(s_t,g)$ defines the advantage function, $\exp_{clip}$ is the exponential advantage weight with a clipped range $(0,M]$, and $\epsilon$ denotes the best-advantage weight to filter the advantage by a threshold. The value function is learned by minimizing the TD-error
     \begin{equation}
        \label{eq:value}
        L_{V_\varphi} = \mathbb{E}_{\tau \sim \mathcal{D}}[(r_t + \gamma V_\varphi(s_{t+1},g) - V_\varphi(s_t,g))^2].
        \end{equation}
     For the experiments we set the discount factor $\gamma$ to be $0.99$, and the clipped upper range $M$ to be $10$. The first $200$ trajectories are collected by executing a random policy to accumulate random trajectories, then the greedy policy $a = \arg \max_a \pi (a|s,g)$ is used.
     \item \textbf{Contrastive GCRL} \citep{eysenbach2022contrastive} uses contrastive learning to train a Q-function relating the current state-action pair and the future states. The loss for the critic function is defined as:
     \begin{equation}
         \label{eq:crl1}
         L=\log \sigma(f(u_{s,a},v_{s^+}))+\log(1-\sigma(f(u_{s,a},v_{s^-}))),
     \end{equation}
     where $u_{s,a}=\phi(s_t,a_t)$ is the representation of the state-action pair, and $v_{s}=\varphi(s_f)$ is the representation of the future state. The positive pairs are sampled from the same trajectory, whereas the negative pairs are randomly sampled from different trajectories.
     The policy is learned to choose the action that can maximize the likelihood that the goal states occur in the future:
     \begin{equation}
         \label{eq:crl2}
         \pi(a|s,g) = \argmax_{\pi(a|s,g)} \mathbb{E}[f(u_{s,a},v_g)].
     \end{equation}
\end{itemize}
\section{Implementation Details}
\subsection{Algorithm}
{ We employ a consistent architecture for the neural networks 
representing the parameterised function $p_\theta$ in GCSL-NF.}
This architecture comprises a fully connected neural network with two hidden layers, featuring $[400, 300]$ neurons and utilizing the SiLU activation function. The input layer merges state and goal information, while the output layer is tailored to match the action space's dimensionality. The network utilizes a logistic activation function for its outputs.

Similarly, for the similarity function in GCSL-NF, we adopt an identical neural network architecture. This structure mirrors the previously described architecture, albeit with the output layer reduced to a single neuron. Logistic activation is applied to the output of this network.

The optimizer is set as Adam optimizer with a learning rate of $0.001$, the regularization coefficient $\alpha$ is set to $0.2$, and the threshold $n$ for the similarity function is set to $5$. Our method follows the deterministic greedy policy $\argmax_a \pi (a|s,g)$ from the beginning, while exploration is encouraged by feedback from the similarity function $p_{\varphi}$.
\subsection{Environments}
\label{appendix:env-details}
Table~\ref{tab:env-details} lists the full parameters of the five discrete-action environments used in Section~\ref{sec:environments}.
\begin{table}[htbp!]
\caption{Parameters of the five discrete-action environments. Initial states and goals are sampled uniformly. The action set ``move'' denotes \{up, down, left, right, stay\}; ``move/turn'' denotes \{forward, backward, turn left, turn right, stay\}.}
\label{tab:env-details}
\centering
\begin{tabular}{lccc>{\raggedright\arraybackslash}p{6.5cm}}
\hline
Environment & Obs.\ dim & Action set & Horizon & Notes \\
\hline
Point Mass Navigation & 2 & move & 50 & Obs.\ range $[-1,1]$; step $0.05$ with $\mathcal{N}(0,0.01)$ noise. \\
\quad + Obstacles & 2 & move & 70 & Adds a central circular obstacle (radius $0.4$, centre $(0,0)$). \\
2D 4-room Navigation & 2 & move & 70 & Obs.\ range $[-1.2,1.2]$; passageways of width $0.4$. \\
2D LiDAR Navigation & 64 & move/turn & 50 & Map $200\times200$; $360^\circ$ LiDAR obs.; step $10$ (move), $\frac{\pi}{2}$ (turn); goals are LiDAR readings at poses $(x,y,\theta)$. \\
Object Pushing & 4 & move & 50 & Obs.\ range $[-0.5,0.5]$; puck side length $0.2$; obs./goal cover end-effector and puck coordinates. \\
\hline
\end{tabular}
\end{table}
\section{Experiment Results}
\label{appendix_results}

\begin{table}[ht]
\caption{Results for baseline algorithms and GCSL-NF}
\label{table:results}
\centering
\resizebox{\textwidth}{!}{
\begin{tabular}{l|c|c|c|c|c|c}\hline
                                                                                           & GCSL-NF            & GCSL              & HER DQN           & HER A2C           & W-GCSL            & \makecell{Contrastive \\ GCRL}  \\\hline
\makecell{Point Mass\\ Navigation}                                                                      & 0.085 $\pm$ 0.037  & 0.187 $\pm$ 0.080 & 0.066 $\pm$ 0.019 & 0.158 $\pm$ 0.029 & 0.127 $\pm$ 0.084 & \textbf{0.065 $\pm$ 0.014} \\\hline

\makecell{Point Mass \\Navigation with\\ Obstacles}          & \textbf{0.152 $\pm$ 0.015}  & 0.341 $\pm$ 0.050 & 0.299 $\pm$ 0.031 & 0.481 $\pm$ 0.105 & 0.364 $\pm$ 0.034 & 0.213 $\pm$ 0.038 \\\hline
\makecell{2D 4-Room \\ Navigation}                                                                        & \textbf{0.495 $\pm$ 0.089}  & 0.781 $\pm$ 0.066 & 0.658 $\pm$ 0.075 & 1.153 $\pm$ 0.312 & 0.775 $\pm$ 0.052 & 0.668 $\pm$ 0.070 \\\hline

\makecell{2D LiDAR Navigation \\with Obstacles\\ - Orientation}&\textbf{ 0.555 $\pm$ 0.070}  & 0.749 $\pm$ 0.087 & 1.597 $\pm$ 0.086 & 1.520 $\pm$ 0.099 & 0.654 $\pm$ 0.147 & 1.227 $\pm$ 0.096 \\\hline
   \makecell{2D LiDAR Navigation \\with Obstacles \\- Position}&\textbf{ 0.344 $\pm$ 0.060}  & 0.435 $\pm$ 0.040 & 0.734 $\pm$ 0.030 & 0.879 $\pm$ 0.058 & 0.379 $\pm$ 0.043 & 0.659 $\pm$ 0.033 \\\hline
  \makecell{Object Pushing\\ Puck Position}                   & \textbf{0.078 $\pm$ 0.009}  & 0.306 $\pm$ 0.035 & 0.463 $\pm$ 0.005 & 0.458 $\pm$ 0.008 & 0.334 $\pm$ 0.031 & 0.465 $\pm$ 0.019 \\\hline
            \makecell{Object Pushing\\ Endeffector Position}& \textbf{0.205 $\pm$ 0.016} & 0.384 $\pm$ 0.019 & 0.321 $\pm$ 0.037 & 0.345 $\pm$ 0.112 & 0.445 $\pm$ 0.016 & 0.440 $\pm$ 0.022\\\hline
\end{tabular}}

\end{table}

\begin{table}[ht]
\caption{Results for baseline algorithms and GCSL-NF for continuous environments}
\label{table:results_continuous}
\centering

\resizebox{\textwidth}{!}{
\begin{tabular}{llcccc}
\hline
\multicolumn{2}{l|}{}                                                                                                                & \multicolumn{1}{l|}{GCSL-NF}           & \multicolumn{1}{l|}{GCSL}              & \multicolumn{1}{l|}{HER DDPG}          & Contrastive GCRL  \\ \hline
\multicolumn{1}{l|}{\multirow{2}{*}{\begin{tabular}[c]{@{}l@{}}Car Racing Point Mass\\ Navigation without Obstacles\end{tabular}}} & \multicolumn{1}{l|}{Position}                & \multicolumn{1}{l|}{0.043 $\pm$ 0.019} & \multicolumn{1}{l|}{\textbf{0.037 $\pm$ 0.010}} & \multicolumn{1}{l|}{0.133 $\pm$ 0.222} & 0.173 $\pm$ 0.007 \\ \cline{2-6} 
\multicolumn{1}{l|}{}                                                                                                              & \multicolumn{1}{l|}{Velocity}                & \multicolumn{1}{l|}{\textbf{0.013 $\pm$ 0.008}} & \multicolumn{1}{l|}{0.016 $\pm$ 0.004} & \multicolumn{1}{l|}{0.054 $\pm$ 0.023} & 0.256 $\pm$ 0.008 \\ \hline
\multicolumn{1}{l|}{\multirow{2}{*}{\begin{tabular}[c]{@{}l@{}}Car Racing Point Mass\\ Navigation with 4-Room\end{tabular}}}       & \multicolumn{1}{l|}{Position}                & \multicolumn{1}{l|}{0.387 $\pm$ 0.077} & \multicolumn{1}{l|}{1.267 $\pm$ 0.045} & \multicolumn{1}{l|}{0.457 $\pm$ 0.021} & \textbf{0.296 $\pm$ 0.055} \\ \cline{2-6} 
\multicolumn{1}{l|}{}                                                                                                              & \multicolumn{1}{l|}{Velocity}                & \multicolumn{1}{l|}{0.095 $\pm$ 0.011} & \multicolumn{1}{l|}{0.233 $\pm$ 0.030} & \multicolumn{1}{l|}{\textbf{0.076 $\pm$ 0.004}} & 0.246 $\pm$ 0.009 \\ \hline

\multicolumn{2}{l|}{Fetch Reach}                                                                                                                                                  & \multicolumn{1}{l|}{0.036 $\pm$ 0.003} & \multicolumn{1}{l|}{\textbf{0.013 $\pm$ 0.013}}                  & \multicolumn{1}{l|}{0.079 $\pm$ 0.021}                  &       0.035 $\pm$ 0.002            \\ \hline
\multicolumn{2}{l|}{Fetch Push}                                                                                                                                                  & \multicolumn{1}{l|}{\textbf{0.049 $\pm$ 0.014}} & \multicolumn{1}{l|}{0.165 $\pm$ 0.009}                  & \multicolumn{1}{l|}{0.171 $\pm$ 0.003}                  &      0.060 $\pm$ 0.007             \\ \hline
       
\end{tabular}}
\end{table}

{
\section{Definitions of alternative similarity measures}
\label{appendix:similarity-defs}

This appendix collects the formal definitions of the two alternative similarity measures compared against the trajectory-induced similarity $p_\varphi$ in Section~\ref{sec:distance-choice}. Table~\ref{tab:table1} summarises all three.
}

SimCLR-TT is an effective contrastive learning method for representation learning with temporal neighbourhoods as positive pairs. The loss of SimCLR-TT is defined as:
\begin{equation}
    \label{eq:simclrtt}
    L(\theta) = -\log \frac{\exp (\text{sim}(z_\theta(s), z_\theta(s^+)))}{\sum_{s^\shortminus} [\exp(\text{sim}(z_\theta(s), z_\theta(s^\shortminus)))]},
\end{equation}
where $z_\theta(s)$ denotes the latent representation of state $s$, $z_\theta(s^+)$ and $z_\theta(s^\shortminus)$ are the latent representations of positive and negative samples, respectively, and $\text{sim}(\cdot, \cdot)$ denotes the cosine similarity. 

The successor representation is a method to predict the expected future state occupancy from any given state. Given a state $s$ and a future state $s^\prime$, the successor representation is defined as $M(s,s^\prime) = \mathbb{E}\left[\sum_{t=0}^{\infty}\gamma^t\mathbbm{1}[s_t=s^\prime]\right]$. With the inspiration from the Bellman equation, the successor representation can be expressed in a recursive form and learned by TD-learning:
\begin{equation}
    \label{eq:sr}
    M(s,s^\prime) = \mathbbm{1}[s_t=s^\prime] + \gamma \mathbb{E}[M(s_{t+1}, s^\prime)].
\end{equation}

\begin{table}[!htbp]
    \caption{Definitions of distance measures\label{tab:table1}}
    \centering
    \begin{tabular}{c|c}
    \hline
    Distance Measure & Definition\\
    \hline
    GCSL-NF-$p_\varphi$ & $p_\varphi(s,s')$ (see Section \ref{sec:distance})\\
    \hline
    SimCLR-TT & $p_{TT}(s,s') = (1+\text{sim}(z_\theta(s),z_\theta(s')))/2$\\
    \hline
    \multirow{3}*{Successor Representation} & $p_{SR}(s,s') = \min (M(s,s')/\hat{M}, 1) $  \\
    ~ & $\hat{M}$ is the normalization parameter defined \\
    ~ & as $0.95$-quantiles of a batch \\
    \hline
    \end{tabular}
    \end{table}

\end{document}